\PassOptionsToPackage{table}{xcolor}
\documentclass{article} %
\usepackage{iclr2027_conference,times}
\usepackage{graphicx}
\graphicspath{{img/}}   

\usepackage{amsmath,amsfonts,bm}

\def\eqref#1{equation~\ref{#1}}

\def\1{\bm{1}}

\DeclareMathAlphabet{\mathsfit}{\encodingdefault}{\sfdefault}{m}{sl}
\SetMathAlphabet{\mathsfit}{bold}{\encodingdefault}{\sfdefault}{bx}{n}

\usepackage{enumitem}
\usepackage{graphicx}
\usepackage{booktabs}
\usepackage{tabularx}
\usepackage{wrapfig}
\usepackage{array}
\usepackage[table]{xcolor}
\usepackage{caption}
\usepackage[most]{tcolorbox}
\usepackage{listings}
\usepackage{float}
\usepackage{placeins}
\usepackage{titletoc}
\definecolor{resultheader}{RGB}{226,236,247}
\newcolumntype{Y}{>{\centering\arraybackslash}X}
\usepackage{hyperref}
\hypersetup{hidelinks}
\usepackage{url}
\usepackage{algorithm}
\usepackage{algpseudocode}
\usepackage{amssymb} 
\usepackage{multirow}
\usepackage{tabularx}
\usepackage{graphicx}
\usepackage{array}
\usepackage{makecell}   %
\usepackage[table]{xcolor}
\usepackage{makecell}
\usepackage{booktabs}
\definecolor{lightblue}{rgb}{0.22,0.45,0.70}
\definecolor{resultheader}{HTML}{DCE6F1}
\definecolor{riseblue}{HTML}{2F75B5}
\definecolor{droporange}{HTML}{D97706}
\definecolor{severered}{HTML}{C94B4B}

\newcommand{\tabledrop}[1]{$_{\color{droporange}\downarrow #1}$}
\newcommand{\tableflat}[1]{$_{\color{gray}\rightarrow #1}$}
\newcommand{\resultcell}[1]{#1}
\newcommand{\effectdown}[1]{\textcolor{droporange}{$\downarrow #1$}}
\newcommand{\effectup}[1]{\textcolor{riseblue}{$\uparrow #1$}}
\newcommand{\effectflat}[1]{\textcolor{gray}{#1}}
\newcommand{\severerate}[1]{\textcolor{severered}{#1}}
\newcommand{\policycell}[2]{#1\,{\color{gray}\scriptsize(#2)}}

\definecolor{promptteal}{HTML}{2A7F7F}
\definecolor{promptpurple}{HTML}{7656A7}
\lstdefinestyle{paperprompt}{
  basicstyle=\ttfamily\scriptsize,
  breaklines=true,
  columns=fullflexible,
  keepspaces=true,
  showstringspaces=false,
  morecomment=[l][\color{riseblue}\bfseries]{\#\#},
  emph={STRICT,CRITICAL,VALID,NEVER},
  emphstyle=\color{severered}\bfseries
}
\lstdefinestyle{papercase}{
  basicstyle=\ttfamily\scriptsize,
  breaklines=true,
  columns=fullflexible,
  keepspaces=true,
  showstringspaces=false,
  morecomment=[l][\color{riseblue}\bfseries]{TASK},
  morecomment=[l][\color{promptteal}\bfseries]{MUTATION},
  morecomment=[l][\color{promptpurple}\bfseries]{PERSONA},
  morecomment=[l][\color{severered}\bfseries]{TIPER},
  morecomment=[l][\color{riseblue}\bfseries]{DEFAULT},
  morecomment=[l][\color{promptpurple}\bfseries]{FAILURE},
  morecomment=[l][\color{riseblue}\bfseries]{TEST},
  morecomment=[l][\color{promptteal}\bfseries]{EXPECTED}
}
\tcbset{paperlisting/.style={
  enhanced jigsaw,
  breakable,
  listing only,
  boxrule=0.6pt,
  arc=1mm,
  fonttitle=\bfseries,
  left=2pt,right=2pt,top=2pt,bottom=2pt,
  before skip=5pt,after skip=7pt
}}

\title{Uncovering Robustness Weaknesses in LLM Agents via Persona Evolution with Task-Integrity Constraints}

\author{%
\textbf{Yutao Xie\textsuperscript{1}, Jingbo Yang\textsuperscript{1}, Guanyu Yao\textsuperscript{1},
Xiaoyu Yang\textsuperscript{2}, Duo Ding\textsuperscript{2}, Shiyu Chang\textsuperscript{1}}\\[7pt]
\small\normalfont\textsuperscript{1}University of California, Santa Barbara\\[2pt]
\small\normalfont\textsuperscript{2}Cresta
}

\iclrfinalcopy %
\begin{document}

\maketitle

\renewcommand{\footnoterule}{%
  \kern-3pt
  \hrule width \ifnum\value{page}=1\textwidth\else12pc\fi
  \kern2.6pt
}
\begingroup
\renewcommand{\thefootnote}{}
\makeatletter
\renewcommand{\@makefntext}[1]{\noindent#1}
\makeatother
\footnotetext{\textit{Corresponding author(s): \href{mailto:yutaoxie@ucsb.edu}{yutaoxie@ucsb.edu}}}
\endgroup

\begin{abstract}
Interactive-agent evaluations commonly rely on standardized LLM-based user simulators that exhibit broadly cooperative behavior, potentially overlooking failures that arise under diverse but realistic user interactions. This can lead to overly optimistic estimates of agent robustness and provide limited failure signals for training agents that generalize across users. Existing approaches largely rely on predefined user behaviors, which may miss agent- or domain-specific weaknesses, while failure-oriented search can reduce task success by invalidating the underlying task rather than exposing genuine robustness failures. To fill this gap, we introduce \textsc{Tiper}, a task-integrity-preserving persona evolution method that automatically discovers agent-specific user policies exposing robustness weaknesses while keeping the underlying task valid and solvable. 
\textsc{Tiper} evolves persistent user personas through reflection-guided behavioral mutations and task-wise Pareto selection under a task-integrity constraint. Across four interactive-agent benchmarks, \textsc{Tiper} yields lower held-out avg@3 scores in all 12 GPT-5.4 settings, with observed reductions of 5.8--33.7 percentage points, and in all 28 agent--domain pairs spanning seven agents, with a median observed reduction of 22.9 points. Further analysis yields a behavior-to-failure taxonomy that connects discovered user behaviors to recurring agent failure modes. Together, these results show that \textsc{Tiper} enables more diagnostic robustness evaluation of interactive agents and provides informative failure analysis for future agent robustness improvement.
Code repository: \url{https://github.com/UCSB-NLP-Chang/TIPER}.

\end{abstract}

\section{Introduction}

Recent advances in reasoning and tool use have made LLM agents increasingly capable of completing multi-turn collaborative tasks that require both user interaction and external actions~\citep{taubench,tau2bench,vitabench}. However, existing evaluations of collaborative agents often oversimplify user dynamics by relying on LLM-based user simulators that are primarily instructed to facilitate task completion. Real users, in contrast, vary substantially in how they communicate, provide information, respond to agent requests, and cooperate toward a shared goal. Without explicit behavioral specifications or training, LLM simulators tend to default to broadly helpful and cooperative interaction patterns, failing to capture this diversity~\citep{naous2026flipping}. As a result, current evaluation protocols may systematically underrepresent challenging yet realistic interaction modes and consequently overestimate agent performance. The same bias can also affect training: agents optimized against a narrow set of simulator behaviors may overfit to these interaction patterns and generalize poorly to diverse users~\citep{marl}. These limitations motivate the systematic identification of user policies that reveal weaknesses in agent robustness, enabling more precise evaluation and providing informative interaction patterns for subsequent agent training.

Recent work has begun to evaluate interactive agents with non-collaborative or behaviorally diverse users. NCUser~\citep{NCUser} introduces predefined classes of non-collaborative behavior, while $\tau$-Trait~\citep{tautrait} enables controllable variation along specified user traits. However, two important limitations remain. \textbf{First, challenging user behaviors are largely specified a priori} through hand-crafted policy classes or researcher-selected traits, which may overlook behaviors that reveal weaknesses specific to a particular agent or domain. AgentNoiseBench~\citep{agentnoisebench} moves toward agent-targeted evaluation by searching for behaviors that degrade agent performance, but operates over message-level perturbations from a fixed taxonomy rather than persistent user policies that shape the interaction over multiple turns. \textbf{Second, reduced task success does not necessarily indicate an agent weakness.} Challenging user behaviors can instead make the underlying task infeasible, confounding agent robustness with task validity. This concern is nontrivial in existing benchmarks: under the $\tau^2$-Bench audit standard~\citep{tau2bench}, our audit of 872 $\tau$-Trait conversations finds task-critical user-simulator errors in 244 cases (28.0\%) (Appendix~\ref{app:tautrait_audit}). We therefore require discovered user policies to preserve the validity and feasibility of the underlying task throughout the realized interaction, a constraint we term \emph{task integrity}. Together, these limitations call for user-policy discovery that is both agent-targeted and task-integrity preserving.

\begin{figure}[h]
  \centering
  \includegraphics[width=\linewidth]{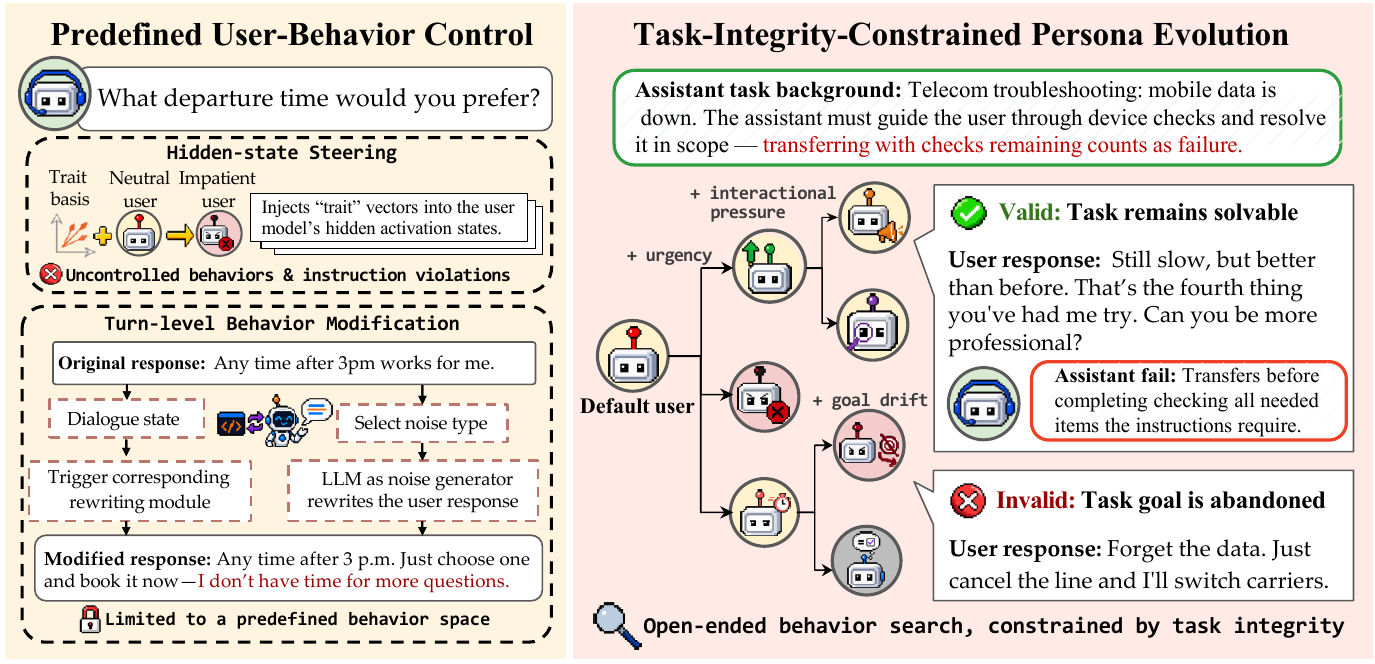}
  \caption{Predefined user-behavior controls versus task-integrity-constrained
           persona evolution.}
  \label{fig:motivation}
\end{figure}

We introduce \textsc{Tiper}, a task-integrity-preserving persona evolution method for automatically identifying user policies that expose weaknesses of a given agent. \textsc{Tiper} iteratively evolves persistent user personas by reflecting on observed interaction failures and proposing behavior-level mutations likely to reveal additional weaknesses. Task-wise Pareto selection preserves complementary failure patterns across tasks, while benchmark-specific validity controls and trajectory audits distinguish agent failures from failures caused by task invalidation. In this way, \textsc{Tiper} searches directly for agent-specific robustness weaknesses while keeping the resulting interactions valid.

Across 12 settings spanning four interactive-agent benchmarks, we observe lower held-out GPT-5.4 task success under \textsc{Tiper}-discovered personas in every setting, with mean differences of 5.8--33.7 percentage points and a median difference of 16.1 points. This pattern extends across model families: across 28 agent--domain pairs covering seven agents, we observe lower task success in every pair, with a median reduction of 22.9 percentage points. In the primary GPT-5.4 evaluation, the pooled task-critical simulator-error rate among exposed failed rollouts is below $2.5\%$, indicating that the discovered failures reflect genuine robustness weaknesses that are missed under standard user simulation. The discovered personas also exhibit stronger behavioral realism than benchmark-default users under human-referenced User-Sim measurements, despite realism not being explicitly optimized during search. Beyond performance degradation, we analyze failures across eight domains to characterize how different user behaviors expose distinct agent failure modes, yielding a behavior-to-failure taxonomy for more diagnostic evaluation and future agent training. Together, these results show that \textsc{Tiper} can automatically uncover robustness weaknesses that current collaborative agent evaluation systematically overlooks.

\section{Problem Formulation}
\label{sec:problem}

Let $\mathcal{X}=\{x_1,\dots,x_N\}$ be a collection of interactive tasks, where
$x_i=(g_i,\mathcal{E}_i,V_i,I_i)$ specifies a user goal, an environment, a task
verifier, and an integrity judge. We fix an evaluated agent $\pi$ and a
user-simulator backbone $\mu$. A persona $p\in\mathcal{P}$ is an
instance-agnostic behavioral instruction that induces the user policy
$\mu_p:=\mu(\cdot\mid p)$.

For persona $p$ on task $x_i$, let $\xi_i^p$ denote the interaction
trajectory induced by $\pi$ and $\mu_p$. The verifier returns
$V_i(\xi_i^p)\in[0,1]$, where $1$ denotes full task success, and the integrity
judge returns $I_i(\xi_i^p)\in\{0,1\}$, where $1$ indicates that the simulated
user preserved the assigned goal, task facts, and environmental constraints.
We define
\begin{equation}
\rho_i(p):=\Pr[I_i(\xi_i^p)=1],
\qquad
\rho(p):=\frac{1}{N}\sum_{i=1}^{N}\rho_i(p),
\qquad
J_i(p):=\mathbb{E}\!\left[
V_i(\xi_i^p)\mid I_i(\xi_i^p)=1
\right].
\label{eq:persona-metrics}
\end{equation}
Here, $\rho(p)$ represents the overall integrity rate of persona $p$ across
the evaluation set. A persona is
\emph{valid on the evaluation set} if
$\rho(p)\geq 1-\varepsilon$,
for a fixed tolerance $\varepsilon\in[0,1)$. Thus, validity is determined by
the aggregate proportion of valid trajectories across tasks. 
Conditioning on $I_i=1$ restricts $J_i(p)$ to trajectories judged to preserve
task integrity.

Let $p_0$ denote the benchmark-default persona. For a persona $p$, define its
exposed performance reduction on task $x_i$ as
$\Delta_i(p)
:=
J_i(p_0)
-J_i(p).
\label{eq:set-score}$
Positive values indicate that the persona lowers agent performance, while
negative values indicate that it improves performance.\footnote{
We assume $\rho_i(p)>0$ for every task $i$, so that the conditional scores $J_i(p)$ are well-defined.
}
We seek a persona that maximizes the average performance reduction while
satisfying the aggregate integrity constraint:
\begin{equation}
\max_{p\in\mathcal{P}} R(p),
\qquad
R(p):=\frac{1}{N}\sum_{i=1}^{N}\Delta_i(p),
\qquad
\text{subject to }\rho(p)\geq 1-\varepsilon.
\label{eq:persona-objective}
\end{equation}
Maximizing $R(p)$ is therefore equivalent to minimizing the persona's mean
task performance.
Neither $\rho_i(p)$ nor $J_i(p)$ is available in closed form, so evaluating a
persona requires rolling out $\pi$ against the induced user policy $\mu_p$.
We therefore compare personas under a fixed evaluation budget. For
\textsc{Tiper}, this budget is determined by the generation limit $G$, the
proposal-attempt limit $M$, the evaluated task sets, and the number of trials
per task.

\section{Task-Integrity-Preserving Persona Evolution}
\label{sec:method}

\begin{figure}[t]
  \centering
  \includegraphics[width=\linewidth]{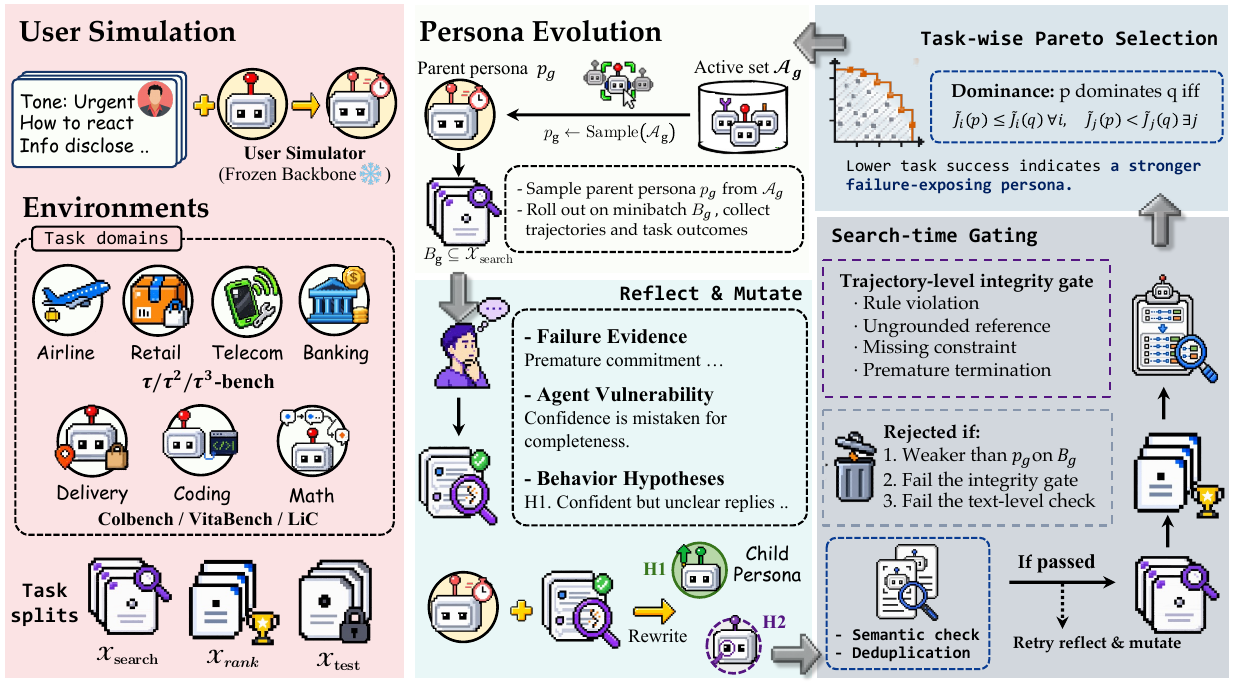}
\caption{
Overview of the task-integrity-preserving persona evolution procedure. Each
generation reflects and mutates a parent using search trajectories. The child undergoes text-level checks and the rollout-based search-time
integrity gate before task-wise Pareto selection updates $\mathcal{A}_{g+1}$.
}
  \label{fig:method}
  \vspace{-12pt}
\end{figure}

To approximate this constrained objective, we perform $G$ generations of
evolutionary search over the natural-language space $\mathcal{P}$. We
initialize $\mathcal{A}_0:=\{p_0\}$. As illustrated in
Figure~\ref{fig:method}, generation $g\in\{0,\ldots,G-1\}$ samples a parent
persona $p_{\mathrm{par}}$ from the active set $\mathcal{A}_g$,
collects trajectories on a minibatch $B_g$, and reflects on them to propose a
child persona. The child undergoes text-level checks and the search-time
integrity gate before task-wise Pareto selection updates the active set.
This section describes the serial search used for $\tau^3$-bench,
Lost-in-Conversation, ColBench, and VitaBench. More details of experiments are described in Appendix~\ref{app:exp-details}.

To separate candidate generation, model selection, and final evaluation, we
partition the tasks into three disjoint subsets.
The search set $\mathcal{X}_{\mathrm{search}}$ is used to generate candidate personas and to perform the
initial parent--child comparisons in each generation. The rank set $\mathcal{X}_{\mathrm{rank}}$ serves
as a validation set for the trajectory-level integrity gate, task-wise Pareto
selection, and final persona selection. The test set $\mathcal{X}_{\mathrm{test}}$ is held out throughout
search and selection and is used only for the final evaluation. Before search, we evaluate $p_0$ on
$\mathcal{X}_{\mathrm{rank}}$ and cache its task-wise scores
$\widetilde{J}_i(p_0)$.

\subsection{Persona Representation and Reflective Proposal}
\label{sec:method:proposal}

\textbf{Persona representation and proposal.}
A persona $p\in\mathcal{P}$ is a domain-aware but instance-agnostic
behavioral instruction. It specifies how the simulated user pursues the
assigned goal without referring to task-specific identifiers, dates, or
instances, allowing the same persona to transfer to unseen tasks in the
domain. Inspired by reflective prompt optimization~\citep{gepa}, the
proposal model reviews a parent persona and its minibatch trajectories,
identifies a behavior that may expose an observed agent weakness, and rewrites
the persona along that direction. 

\textbf{Text-level checks.}
Before rollout, two filters screen each proposal. The first rejects personas
that are unclear, instance-specific, or inconsistent with the assigned task.
The second rejects near-duplicates using lexical and embedding similarity
against the proposal history. Together, these filters prevent invalid or
redundant candidates from consuming rollout budget and keep the search focused
on transferable behavioral changes. Proposals that fail either check are
revised within a retry loop of at most $M$ attempts, whereas those that pass
enter the search-time integrity gate. Both text-check feedback and
gate-rejection reasons are supplied to the proposal model, allowing subsequent
attempts to address the identified issue while preserving the intended mutation
direction.

\subsection{Search-Time Integrity Gating}
\label{sec:search-time-gating}

For each generation, we first compare the child persona with its parent on the
search minibatch $B_g$. Let $\widetilde{J}_i(p)$ denote the 
performance of $p$ on task $x_i$, and define
$\widetilde{J}_{B_g}(p)
:=
\frac{1}{|B_g|}
\sum_{x_i\in B_g}\widetilde{J}_i(p)$. 

A child that yields a lower minibatch score than its parent is then
evaluated on $\mathcal{X}_{\mathrm{rank}}$. Let
$\widetilde{I}_i(\xi)\in\{0,1\}$ indicate whether rollout $\xi$ on task $x_i$
passes the search-time integrity check. We define
$\widetilde{\rho}_{\mathrm{fail}}(p)$ as the fraction of failure rollouts on $\mathcal{X}_{\mathrm{rank}}$ that pass this integrity check. A child $\tilde p$ passes the search-time integrity
gate if
$\widetilde{J}_{B_g}(\tilde p)
<
\widetilde{J}_{B_g}(p_{\mathrm{par}})$ and $
\widetilde{\rho}_{\mathrm{fail}}(\tilde p)\geq 1-\gamma$, 
where $\gamma\in[0,1]$ is the maximum allowed integrity-violation rate among
verifier-failed rollouts. The first condition favors children that expose
more agent failures, while the second prevents the search from favoring
failures caused by violations of the simulated user's behavior.
This search-time gate is an empirical filter and does not certify the formal
validity constraint defined by $\rho(p)$. Rejected children leave the active
set unchanged, while gate-passing children proceed to task-wise Pareto
selection.

\subsection{Task-Wise Pareto Selection}
\label{sec:pareto-selection}

For candidates that pass the integrity gate, we determine Pareto dominance
from their task-wise rank-set scores. Since a lower
score indicates stronger failure exposure, persona $p$ dominates persona $q$
if $\widetilde{J}_i(p)\leq\widetilde{J}_i(q)$ for every
$x_i\in\mathcal{X}_{\mathrm{rank}}$, with strict inequality for at least one
task.

Let $\operatorname{ND}_{\widetilde{J}}(\mathcal{S})$ denote the nondominated
members of $\mathcal{S}$ under this relation. Let
$\operatorname{KeepK}(\mathcal{S},K)$ retain all members when
$|\mathcal{S}|\leq K$ and otherwise retain the
$K$ members with the lowest mean rank-set score. For a gate-passing child
$\tilde p$, we update
$\mathcal{A}_{g+1}:=
\operatorname{KeepK}(\operatorname{ND}_{\widetilde{J}}
(\mathcal{A}_g\cup\{\tilde p\}),K)$.
If no proposal attempt passes the checks and gate, then
$\mathcal{A}_{g+1}:=\mathcal{A}_g$.
Task-wise selection retains personas that expose complementary weaknesses
during evolution. Parents for the next generation are sampled from the updated
active set. After $G$ generations, we select one persona from $\mathcal{A}_G$
using the rank set:
\begin{equation}
\hat p:=\arg\min_{p\in\mathcal{A}_G}
\frac{1}{|\mathcal{X}_{\mathrm{rank}}|}
\sum_{x_i\in\mathcal{X}_{\mathrm{rank}}}\widetilde{J}_i(p).
\label{eq:final-persona-selection}
\end{equation}
We fix $\hat p$ before test evaluation and report all test metrics for this
persona.

\section{Experiments}
\label{sec:exp}
We evaluate \textsc{Tiper} through four research questions:
\begin{itemize}[leftmargin=*, itemsep=1pt, topsep=1pt]
    \item \textbf{(Q1) Failure-revealing persona evolution.}
    Can \textsc{Tiper} discover substantially more failures across held-out test tasks than the default and alternative policies?
    \item \textbf{(Q2) Robustness across agents and components.}
    How does failure exposure with \textsc{Tiper} vary across evaluated agents, mutation models,
    and simulator backbones?
    \item \textbf{(Q3) Integrity and behavioral alignment.}
    How well do the generated trajectories preserve task integrity and align
    with human-user references?
    \item \textbf{(Q4) Behavior-to-failure characterization.}
    How do task-preserving user behaviors vary across benchmarks, and what agent
    failure modes are associated with them?
\end{itemize}

\subsection{Experimental Setup}
\label{sec:setup}

\textbf{Benchmarks and agents.}
We evaluate \textsc{Tiper} with GPT-5.4~\citep{GPT-5.4} on 12 settings drawn
from four interactive-agent benchmarks: $\tau^3$-bench~\citep{tau2bench, tauknowledge},
Lost-in-Conversation~\citep{lic}, ColBench~\citep{sweetrl}, and VitaBench~\citep{vitabench}. To study generalization across
agents, we rerun the persona search for six additional proprietary and
open-weight agents: GLM-5.2~\citep{GLM-5.2},
Claude-Sonnet-5~\citep{Claude-Sonnet-5},
DS-V4-Flash~\citep{DS-V4-Flash},
MiniMax-M3~\citep{MiniMax-M3},
Qwen3.7-Plus~\citep{Qwen3.7-Plus}, and
MUSE-Glimmer~\citep{MUSE-Glimmer}. This cross-agent study covers four
representative domains: Telecom, Math, Coding, and Delivery. Unless otherwise
specified, we use GPT-5.4 as the simulator backbone and GPT-5.5 as the mutation
model. GPT-5.2 and GLM-5.2 serve as the judges. More
implementation details are provided in Appendix~\ref{app:exp-details}.

\textbf{Baselines.}
We compare \textsc{Tiper} with the benchmark-default user simulator, the fixed
user programs from NCUser~\citep{NCUser}, and an objective-matched variant of
PPol~\citep{ppol}. The PPol variant uses the same agent-difficulty objective,
task partitions, and search budget as \textsc{Tiper}. All methods are evaluated
on the same test tasks and undergo the identical trajectory-level integrity audit
described in Appendix~\ref{app:integrity-evaluation}.

\textbf{Evaluation protocol and metrics.}
Each \textsc{Tiper} candidate that reaches rank-set evaluation is run twice per
task. We select the persona with the lowest mean rank-set score and fix it
before test evaluation. We then conduct three independent test rollouts per
task.

After integrity filtering, we report valid avg@3, defined as the mean task
score over the valid rollouts among the three test trials. We also report
$\mathrm{pass}^k$, defined among tasks with three valid rollouts as the
fraction with at least $k$ fully successful rollouts, where
$k\in\{1,2,3\}$. For ColBench, valid avg@3 is the mean fraction of unit tests
passed, and a rollout is fully successful only if it passes all unit
tests.

We define failure exposure as the reduction in task score from the default
condition to the persona condition after integrity filtering. In the following experiments, every test task has at least one integrity-valid rollout in each reported condition.
Policy comparisons additionally report the failure-conditioned task-critical
error rate under an adapted $\tau^2$-Bench rubric~\citep{tau2bench}. The User-Sim
Index measures behavioral alignment between trajectories induced by
\textsc{Tiper} and human-user reference trajectories~\citep{sim2real}.
Because integrity filtering can change the tasks eligible for
$\mathrm{pass}^k$, we interpret these metrics descriptively.
Appendix~\ref{app:common-valid-sensitivity} additionally examines avg@3
sensitivity to common-valid-task filtering for the four fixed-policy
comparisons on Airline.

\subsection{Main Results}
\textsc{Tiper} consistently exposes additional agent failures on held-out
tasks. As shown in Table~\ref{tab:benchmark-domain-degradation}, the selected
personas reduce both avg@3 and $\mathrm{pass}^3$ in all 12 settings, with
median reductions of 16.1 and 15.0 percentage points, respectively. The
largest avg@3 reductions occur in Math, Telecom, and Delivery, at 33.7, 25.3,
and 25.0 points, with corresponding $\mathrm{pass}^3$ reductions of 54.1,
35.2, and 22.5 points. Because persona selection does not use the test tasks,
these results indicate that the discovered behaviors transfer to unseen tasks
within each domain.

\begin{table*}[t]
\centering
\caption{
Held-out failure exposure across benchmarks for GPT-5.4. We compare the
benchmark-default user simulator with the rank-selected \textsc{Tiper} persona
on 12 test settings. All values are percentages. The orange numbers indicate the decrease from the corresponding default
scores in percentage points. Differences are computed from the displayed rounded scores.
}
\label{tab:benchmark-domain-degradation}
\setlength{\tabcolsep}{8pt}
\renewcommand{\arraystretch}{1.1}
\rowcolors{3}{white}{gray!10}
\resizebox{\textwidth}{!}{%
\begin{tabular}{ll|cc|cc|cc|cc}
\Xhline{1.2pt}
\rowcolor{resultheader}
\multicolumn{2}{c|}{\cellcolor{resultheader}\textbf{Setting}} &
\multicolumn{2}{c|}{\cellcolor{resultheader}\textbf{avg@3}} &
\multicolumn{2}{c|}{\cellcolor{resultheader}\textbf{pass$^1$}} &
\multicolumn{2}{c|}{\cellcolor{resultheader}\textbf{pass$^2$}} &
\multicolumn{2}{c}{\cellcolor{resultheader}\textbf{pass$^3$}} \\
\hline
\rowcolor{resultheader}
\textbf{Benchmark} & \textbf{Domain} &
\textbf{default} & \textbf{\textsc{Tiper}} &
\textbf{default} & \textbf{\textsc{Tiper}} &
\textbf{default} & \textbf{\textsc{Tiper}} &
\textbf{default} & \textbf{\textsc{Tiper}} \\
\Xhline{1.2pt}
& Telecom  & 81.5 & \resultcell{56.2\tabledrop{25.3}} & 94.4 & \resultcell{77.8\tabledrop{16.6}} & 79.6 & \resultcell{55.6\tabledrop{24.0}} & 70.4 & \resultcell{35.2\tabledrop{35.2}} \\
& Banking  & 33.3 & \resultcell{27.5\tabledrop{5.8}}  & 45.6 & \resultcell{43.9\tabledrop{1.7}} & 35.1 & \resultcell{22.8\tabledrop{12.3}} & 19.3 & \resultcell{15.8\tabledrop{3.5}} \\
& Retail   & 78.4 & \resultcell{71.6\tabledrop{6.8}}  & 92.6 & \resultcell{83.3\tabledrop{9.3}} & 85.2 & \resultcell{75.9\tabledrop{9.3}} & 57.4 & \resultcell{55.6\tabledrop{1.8}} \\
\multirow{-4}{*}{$\tau^3$-bench~{\scriptsize\citep{tau2bench}}}
& Airline  & 86.7 & \resultcell{76.7\tabledrop{10.0}} & 92.0 & \resultcell{82.0\tabledrop{10.0}} & 90.0 & \resultcell{78.0\tabledrop{12.0}} & 78.0 & \resultcell{70.0\tabledrop{8.0}} \\
\hline
& Math     & 89.9 & \resultcell{56.2\tabledrop{33.7}} & 97.7 & \resultcell{82.5\tabledrop{15.2}} & 93.0 & \resultcell{60.0\tabledrop{33.0}} & 79.1 & \resultcell{25.0\tabledrop{54.1}} \\
& Database & 65.2 & \resultcell{50.4\tabledrop{14.8}} & 76.6 & \resultcell{68.1\tabledrop{8.5}} & 68.1 & \resultcell{46.8\tabledrop{21.3}} & 51.1 & \resultcell{36.2\tabledrop{14.9}} \\
\multirow{-3}{*}{Lost-in-Conv.~{\scriptsize\citep{lic}}}
& Actions  & 52.6 & \resultcell{40.0\tabledrop{12.6}} & 55.6 & \resultcell{51.1\tabledrop{4.5}} & 53.3 & \resultcell{40.0\tabledrop{13.3}} & 48.9 & \resultcell{28.9\tabledrop{20.0}} \\
\hline
ColBench~{\scriptsize\citep{sweetrl}}
& Coding   & 44.6 & \resultcell{27.2\tabledrop{17.4}} & 32.0 & \resultcell{12.0\tabledrop{20.0}} & 20.0 & \resultcell{6.0\tabledrop{14.0}} & 14.0 & \resultcell{2.0\tabledrop{12.0}} \\
\hline
& Delivery     & 66.7 & \resultcell{41.7\tabledrop{25.0}} & 85.0 & \resultcell{62.5\tabledrop{22.5}} & 72.5 & \resultcell{42.5\tabledrop{30.0}} & 42.5 & \resultcell{20.0\tabledrop{22.5}} \\
& In-store     & 76.7 & \resultcell{66.7\tabledrop{10.0}} & 92.5 & \resultcell{82.5\tabledrop{10.0}} & 77.5 & \resultcell{72.5\tabledrop{5.0}} & 60.0 & \resultcell{45.0\tabledrop{15.0}} \\
& OTA          & 51.7 & \resultcell{34.2\tabledrop{17.5}} & 82.5 & \resultcell{55.0\tabledrop{27.5}} & 45.0 & \resultcell{27.5\tabledrop{17.5}} & 27.5 & \resultcell{20.0\tabledrop{7.5}} \\
\multirow{-4}{*}{VitaBench~{\scriptsize\citep{vitabench}}}
& Cross-domain & 54.2 & \resultcell{34.2\tabledrop{20.0}} & 77.5 & \resultcell{62.5\tabledrop{15.0}} & 57.5 & \resultcell{30.0\tabledrop{27.5}} & 27.5 & \resultcell{10.0\tabledrop{17.5}} \\
\Xhline{1.2pt}
\end{tabular}
}
\vspace{-10pt}
\end{table*}

\textbf{Cross-agent generalization.}
\textsc{Tiper} consistently exposes failures across different agent models. Across 28 combinations of seven agents and four domains in Table~\ref{tab:main_results},
avg@3 decreases in every setting and $\mathrm{pass}^3$ decreases in 25, with median reductions of 22.9 and
20.4 percentage points, respectively. The
effect is particularly consistent in Telecom, where avg@3 decreases by
13.0--39.5 points and $\mathrm{pass}^3$ by 18.6--44.5 points across all seven
agents. The magnitude varies more in Math, with avg@3 reductions ranging from
0.8 points for GLM-5.2 to 33.7 points for GPT-5.4.

\textbf{Baseline comparison.}
To assess the value of agent-conditioned persona search, we compare
\textsc{Tiper} with fixed behavior programs and objective-matched PPol across
four domains and two agents (Table~\ref{tab:policy-comparison}). Across all
eight agent--domain settings, \textsc{Tiper} produces the largest avg@3
reduction, exceeding the strongest alternative by a median of 7.5 percentage
points. Overall, \textsc{Tiper} exposes
more agent failures than either fixed behaviors or objective-matched PPol.

\begin{table*}[t]
\centering
\caption{
Cross-agent held-out failure exposure on four representative domains. For each
agent and domain, we rerun the same \textsc{Tiper} search. All values are percentages.
The orange numbers indicate the decrease
from the corresponding default scores in percentage points. Differences are computed from the displayed rounded scores.
}
\label{tab:main_results}
\setlength{\tabcolsep}{8pt}
\renewcommand{\arraystretch}{1.1}
\rowcolors{3}{white}{gray!10}
\resizebox{\textwidth}{!}{%
\begin{tabular}{ll|cc|cc|cc|cc}
\rowcolor{resultheader}
\multicolumn{2}{c|}{\cellcolor{resultheader}\textbf{Setting}} &
\multicolumn{2}{c|}{\cellcolor{resultheader}\textbf{avg@3}} &
\multicolumn{2}{c|}{\cellcolor{resultheader}\textbf{pass$^1$}} &
\multicolumn{2}{c|}{\cellcolor{resultheader}\textbf{pass$^2$}} &
\multicolumn{2}{c}{\cellcolor{resultheader}\textbf{pass$^3$}} \\
\hline
\rowcolor{resultheader}
\textbf{Domain} & \textbf{Agent model} &
\textbf{default} & \textbf{\textsc{Tiper}} &
\textbf{default} & \textbf{\textsc{Tiper}} &
\textbf{default} & \textbf{\textsc{Tiper}} &
\textbf{default} & \textbf{\textsc{Tiper}} \\
\Xhline{1.2pt}
& GPT-5.4              & 81.5 & \resultcell{56.2\tabledrop{25.3}} & 94.4 & \resultcell{77.8\tabledrop{16.6}} & 79.6 & \resultcell{55.6\tabledrop{24.0}} & 70.4 & \resultcell{35.2\tabledrop{35.2}} \\
& GLM-5.2              & 95.1 & \resultcell{69.8\tabledrop{25.3}} & 100.0 & \resultcell{83.3\tabledrop{16.7}} & 100.0 & \resultcell{74.1\tabledrop{25.9}} & 85.2 & \resultcell{51.9\tabledrop{33.3}} \\
& Claude-Sonnet-5      & 69.8 & \resultcell{44.4\tabledrop{25.4}} & 83.3 & \resultcell{51.9\tabledrop{31.4}} & 74.1 & \resultcell{48.1\tabledrop{26.0}} & 51.9 & \resultcell{33.3\tabledrop{18.6}} \\
& DS-V4-Flash    & 90.7 & \resultcell{77.7\tabledrop{13.0}} & 98.1 & \resultcell{94.4\tabledrop{3.7}} & 96.3 & \resultcell{83.3\tabledrop{13.0}} & 77.8 & \resultcell{55.6\tabledrop{22.2}} \\
& MiniMax-M3           & 90.1 & \resultcell{53.1\tabledrop{37.0}} & 100.0 & \resultcell{61.1\tabledrop{38.9}} & 94.4 & \resultcell{55.6\tabledrop{38.8}} & 75.9 & \resultcell{42.6\tabledrop{33.3}} \\
& Qwen3.7-Plus         & 94.4 & \resultcell{54.9\tabledrop{39.5}} & 100.0 & \resultcell{68.5\tabledrop{31.5}} & 98.1 & \resultcell{55.6\tabledrop{42.5}} & 85.2 & \resultcell{40.7\tabledrop{44.5}} \\
\multirow{-7}{*}{Telecom~{\scriptsize\citep{tau2bench}}}
& MUSE-Glimmer         & 80.9 & \resultcell{49.4\tabledrop{31.5}} & 96.3 & \resultcell{70.4\tabledrop{25.9}} & 79.6 & \resultcell{53.7\tabledrop{25.9}} & 66.7 & \resultcell{24.1\tabledrop{42.6}} \\
\hline
& GPT-5.4              & 44.6 & \resultcell{27.2\tabledrop{17.4}} & 32.0 & \resultcell{12.0\tabledrop{20.0}} & 20.0 & \resultcell{6.0\tabledrop{14.0}} & 14.0 & \resultcell{2.0\tabledrop{12.0}} \\
& GLM-5.2              & 70.2 & \resultcell{51.6\tabledrop{18.6}} & 56.0 & \resultcell{26.0\tabledrop{30.0}} & 34.0 & \resultcell{20.0\tabledrop{14.0}} & 16.0 & \resultcell{2.0\tabledrop{14.0}} \\
& Claude-Sonnet-5      & 77.8 & \resultcell{50.0\tabledrop{27.8}} & 62.0 & \resultcell{32.0\tabledrop{30.0}} & 38.0 & \resultcell{12.0\tabledrop{26.0}} & 18.0 & \resultcell{6.0\tabledrop{12.0}} \\
& DS-V4-Flash    & 67.6 & \resultcell{40.0\tabledrop{27.6}} & 54.0 & \resultcell{24.0\tabledrop{30.0}} & 28.0 & \resultcell{10.0\tabledrop{18.0}} & 6.0 & \resultcell{6.0\tableflat{0.0}} \\
& MiniMax-M3           & 76.6 & \resultcell{72.8\tabledrop{3.8}}  & 68.0 & \resultcell{58.0\tabledrop{10.0}} & 30.0 & \resultcell{18.0\tabledrop{12.0}} & 12.0 & \resultcell{6.0\tabledrop{6.0}} \\
& Qwen3.7-Plus         & 45.8 & \resultcell{38.2\tabledrop{7.6}} & 36.0 & \resultcell{18.0\tabledrop{18.0}} & 24.0 & \resultcell{12.0\tabledrop{12.0}} & 8.0 & \resultcell{8.0\tableflat{0.0}} \\
\multirow{-7}{*}{Coding~{\scriptsize\citep{sweetrl}}}
& MUSE-Glimmer         & 78.8 & \resultcell{60.6\tabledrop{18.2}} & 68.0 & \resultcell{42.0\tabledrop{26.0}} & 46.0 & \resultcell{24.0\tabledrop{22.0}} & 24.0 & \resultcell{16.0\tabledrop{8.0}} \\
\hline
& GPT-5.4              & 89.9 & \resultcell{56.2\tabledrop{33.7}} & 97.7 & \resultcell{82.5\tabledrop{15.2}} & 93.0 & \resultcell{60.0\tabledrop{33.0}} & 79.1 & \resultcell{25.0\tabledrop{54.1}} \\
& GLM-5.2              & 84.5 & \resultcell{83.7\tabledrop{0.8}}  & 95.3 & \resultcell{95.3\tableflat{0.0}} & 86.0 & \resultcell{83.7\tabledrop{2.3}} & 72.1 & \resultcell{72.1\tableflat{0.0}} \\
& Claude-Sonnet-5      & 89.9 & \resultcell{71.3\tabledrop{18.6}} & 95.3 & \resultcell{95.3\tableflat{0.0}} & 93.0 & \resultcell{69.8\tabledrop{23.2}} & 81.4 & \resultcell{48.8\tabledrop{32.6}} \\
& DS-V4-Flash    & 82.2 & \resultcell{74.4\tabledrop{7.8}}  & 95.3 & \resultcell{90.7\tabledrop{4.6}} & 86.0 & \resultcell{76.7\tabledrop{9.3}} & 65.1 & \resultcell{55.8\tabledrop{9.3}} \\
& MiniMax-M3           & 81.4 & \resultcell{73.6\tabledrop{7.8}}  & 95.3 & \resultcell{93.0\tabledrop{2.3}} & 86.0 & \resultcell{72.1\tabledrop{13.9}} & 62.8 & \resultcell{55.8\tabledrop{7.0}} \\
& Qwen3.7-Plus         & 49.6 & \resultcell{35.7\tabledrop{13.9}} & 79.1 & \resultcell{65.1\tabledrop{14.0}} & 53.5 & \resultcell{32.6\tabledrop{20.9}} & 16.3 & \resultcell{9.3\tabledrop{7.0}} \\
\multirow{-7}{*}{Math~{\scriptsize\citep{lic}}}
& MUSE-Glimmer         & 50.4 & \resultcell{39.5\tabledrop{10.9}} & 79.1 & \resultcell{62.8\tabledrop{16.3}} & 46.5 & \resultcell{37.2\tabledrop{9.3}} & 25.6 & \resultcell{18.6\tabledrop{7.0}} \\
\hline
& GPT-5.4              & 66.7 & \resultcell{41.7\tabledrop{25.0}} & 85.0 & \resultcell{62.5\tabledrop{22.5}} & 72.5 & \resultcell{42.5\tabledrop{30.0}} & 42.5 & \resultcell{20.0\tabledrop{22.5}} \\
& GLM-5.2              & 62.5 & \resultcell{55.0\tabledrop{7.5}}  & 82.5 & \resultcell{75.0\tabledrop{7.5}} & 62.5 & \resultcell{62.5\tableflat{0.0}} & 42.5 & \resultcell{27.5\tabledrop{15.0}} \\
& Claude-Sonnet-5      & 64.2 & \resultcell{39.2\tabledrop{25.0}} & 82.5 & \resultcell{67.5\tabledrop{15.0}} & 72.5 & \resultcell{35.0\tabledrop{37.5}} & 37.5 & \resultcell{15.0\tabledrop{22.5}} \\
& DS-V4-Flash    & 68.3 & \resultcell{46.7\tabledrop{21.6}} & 87.5 & \resultcell{70.0\tabledrop{17.5}} & 70.0 & \resultcell{47.5\tabledrop{22.5}} & 47.5 & \resultcell{22.5\tabledrop{25.0}} \\
& MiniMax-M3           & 60.0 & \resultcell{32.5\tabledrop{27.5}} & 82.5 & \resultcell{62.5\tabledrop{20.0}} & 67.5 & \resultcell{27.5\tabledrop{40.0}} & 30.0 & \resultcell{7.5\tabledrop{22.5}} \\
& Qwen3.7-Plus         & 70.0 & \resultcell{45.8\tabledrop{24.2}} & 87.5 & \resultcell{75.0\tabledrop{12.5}} & 72.5 & \resultcell{42.5\tabledrop{30.0}} & 50.0 & \resultcell{20.0\tabledrop{30.0}} \\
\multirow{-7}{*}{Delivery~{\scriptsize\citep{vitabench}}}
& MUSE-Glimmer         & 54.2 & \resultcell{28.3\tabledrop{25.9}} & 77.5 & \resultcell{55.0\tabledrop{22.5}} & 57.5 & \resultcell{25.0\tabledrop{32.5}} & 27.5 & \resultcell{5.0\tabledrop{22.5}} \\
\Xhline{1.2pt}
\end{tabular}
}
\end{table*}

\subsection{Integrity and Behavioral Alignment}

\textbf{Integrity.}
Integrity is assessed at the trajectory level, and trajectories judged invalid
are excluded before task scores are aggregated. Among failed rollouts on
held-out tasks, Table~\ref{tab:policy-comparison} reports the fraction
containing a task-critical user-simulator error. This rate is at most 2.5\%
for \textsc{Tiper}, compared with up to 21.4\% for the alternative policies.
Together, these results support interpreting \textsc{Tiper}'s larger score
reductions as exposure of agent weaknesses instead of artifacts of
task-integrity violations. A blinded human audit of 150 automatically accepted TIPER failure trajectories found 148 task-valid cases, of which 143 had an identifiable agent error (Appendix~\ref{app:human-integrity-audit}).

\textbf{Behavioral alignment.}
The left panels of Figure~\ref{fig:behavioral-plausibility-and-behaviors}
compare our user-policy trajectories with Sim2Real references along
communication style, information disclosure, clarification, and reactions to
agent errors. ECE measures differences in agent success across task-difficulty
levels. We plot $(1-\mathrm{ECE})\times100$, so farther outward indicates
closer alignment, while $\mathrm{USI}$ aggregates all five metrics. On Airline
and Retail \textsc{Tiper} achieves higher aggregate USI scores than the benchmark-default
simulator in both domains, while remaining below the human-reference scores
(Table~\ref{tab:policy-usi}).

\begin{table*}[t]
\centering
\caption{
Policy comparison across four domains and two agents. Entries show changes in
avg@3 from the default setting in percentage points
($\downarrow$ lower, $\uparrow$ higher), followed in parentheses by the
percentage of failed rollouts with task-critical simulator errors under the
$\tau^2$-Bench rubric~\citep{tau2bench}. Error rates above 5\% are shown in red.
}
\label{tab:policy-comparison}
\setlength{\tabcolsep}{4pt}
\renewcommand{\arraystretch}{1.1}
\rowcolors{3}{white}{gray!10}
\resizebox{\textwidth}{!}{%
\begin{tabular}{l|cc|cc|cc|cc}
\Xhline{1.2pt}
\rowcolor{resultheader}
\textbf{Domain} & \multicolumn{2}{c|}{\cellcolor{resultheader}\textbf{Telecom}} &
\multicolumn{2}{c|}{\cellcolor{resultheader}\textbf{Airline}} &
\multicolumn{2}{c|}{\cellcolor{resultheader}\textbf{Retail}} &
\multicolumn{2}{c}{\cellcolor{resultheader}\textbf{Delivery}} \\
\hline
\rowcolor{resultheader}
\textbf{Policy} & \textbf{GPT-5.4} & \textbf{DS-V4-Flash} & \textbf{GPT-5.4} &
\textbf{DS-V4-Flash} & \textbf{GPT-5.4} & \textbf{DS-V4-Flash} & \textbf{GPT-5.4} &
\textbf{DS-V4-Flash} \\
\Xhline{1.2pt}

Default setting &
\policycell{---}{0.0\%} & \policycell{---}{0.0\%} &
\policycell{---}{0.0\%} & \policycell{---}{\severerate{16.7\%}} &
\policycell{---}{3.1\%} & \policycell{---}{\severerate{8.3\%}} &
\policycell{---}{0.0\%} & \policycell{---}{0.0\%} \\

Emotional &
\policycell{\effectdown{6.2}}{2.5\%} & \policycell{\effectup{0.6}}{\severerate{21.4\%}} &
\policycell{\effectdown{9.0}}{\severerate{10.0\%}} & \policycell{\effectflat{0.0}}{0.0\%} &
\policycell{\effectflat{0.0}}{2.9\%} & \policycell{\effectup{0.6}}{0.0\%} &
\policycell{\effectdown{9.2}}{2.0\%} & \policycell{\effectup{3.3}}{2.9\%} \\

Tangential &
\policycell{\effectdown{14.2}}{1.9\%} & \policycell{\effectdown{10.2}}{0.0\%} &
\policycell{\effectdown{2.0}}{0.0\%} & \policycell{\effectup{6.7}}{0.0\%} &
\policycell{\effectup{6.2}}{4.0\%} & \policycell{\effectdown{0.6}}{0.0\%} &
\policycell{\effectdown{5.8}}{4.3\%} & \policycell{\effectdown{3.3}}{0.0\%} \\

Fragment &
\policycell{\effectdown{5.6}}{0.0\%} & \policycell{\effectdown{3.1}}{0.0\%} &
\policycell{\effectdown{1.3}}{0.0\%} & \policycell{\effectup{3.3}}{\severerate{6.7\%}} &
\policycell{\effectup{4.9}}{3.7\%} & \policycell{\effectdown{1.2}}{3.8\%} &
\policycell{\effectdown{5.8}}{2.1\%} & \policycell{\effectdown{7.5}}{2.1\%} \\

Unavailable &
\policycell{\effectup{2.5}}{0.0\%} & \policycell{\effectup{3.7}}{0.0\%} &
\policycell{\effectdown{5.3}}{0.0\%} & \policycell{\effectup{10.0}}{0.0\%} &
\policycell{\effectup{4.3}}{3.6\%} & \policycell{\effectflat{0.0}}{4.2\%} &
\policycell{\effectdown{9.2}}{\severerate{7.8\%}} & \policycell{\effectdown{0.8}}{2.6\%} \\

PPol &
\policycell{\effectdown{1.9}}{\severerate{14.3\%}} & \policycell{\effectup{4.3}}{0.0\%} &
\policycell{\effectdown{1.0}}{\severerate{7.7\%}} & \policycell{\effectup{1.0}}{0.0\%} &
\policycell{\effectdown{4.9}}{2.3\%} & \policycell{\effectup{0.6}}{0.0\%} &
\policycell{\effectdown{10.0}}{0.0\%} & \policycell{\effectdown{13.3}}{0.0\%} \\

\hline
\textbf{\textsc{Tiper}} &
\policycell{\textbf{\effectdown{25.3}}}{\textbf{0.0\%}} & \policycell{\textbf{\effectdown{13.0}}}{\textbf{0.0\%}} &
\policycell{\textbf{\effectdown{10.0}}}{\textbf{0.0\%}} & \policycell{\textbf{\effectdown{6.7}}}{\textbf{0.0\%}} &
\policycell{\textbf{\effectdown{6.8}}}{\textbf{1.6\%}} & \policycell{\textbf{\effectdown{9.9}}}{\textbf{2.5\%}} &
\policycell{\textbf{\effectdown{25.0}}}{\textbf{0.0\%}} & \policycell{\textbf{\effectdown{21.6}}}{\textbf{0.0\%}} \\

\Xhline{1.2pt}
\end{tabular}
}
\end{table*}

\begin{table*}[t]
\centering
\small
\caption{\textbf{Search ablation and component sensitivity.}
Percentage-point changes from default; more negative values indicate stronger
failure exposure. Panel (a) removes the integrity gate or post-proposal persona
check; panels (b) and (c) vary the mutation model and simulator backbone. In panel (c), each simulator backbone is compared against its own
default-persona baseline.}
\label{tab:ablation-sensitivity}

\setlength{\tabcolsep}{3pt}
\renewcommand{\arraystretch}{1.1}

\begin{minipage}[t]{0.35\textwidth}
  \vspace{0pt}
  \centering
  \rowcolors{4}{white}{gray!10}

  \resizebox{\linewidth}{!}{%
  \begin{tabular}{l|cc|cc}
    \Xhline{1.2pt}

    \rowcolor{resultheader}
    \multicolumn{5}{c}{
      \cellcolor{resultheader}\textbf{(a) Search component ablation (Telecom)}
    } \\
    \hline

    \rowcolor{resultheader}
    &
    \multicolumn{2}{c|}{
      \cellcolor{resultheader}\textbf{GPT-5.4}
    } &
    \multicolumn{2}{c}{
      \cellcolor{resultheader}\textbf{DS-V4-Flash}
    } \\

    \rowcolor{resultheader}
    \multirow{-2}{*}{\textbf{Method}} &
    \textbf{avg@3} & \textbf{pass$^3$} &
    \textbf{avg@3} & \textbf{pass$^3$} \\
    \Xhline{1.2pt}

    \textbf{\textsc{Tiper}} &
    $\mathbf{-25.3}$ & $\mathbf{-35.2}$ &
    $\mathbf{-13.0}$ & $\mathbf{-22.2}$ \\

    w/o integrity gate &
    $-10.5$ & $-14.8$ &
    $-8.0$ & $-14.1$ \\
    
    w/o persona check &
    $-4.3$ & $-8.1$ &
    $-4.9$ & $-8.5$ \\
    \Xhline{1.2pt}
  \end{tabular}%
  }
\end{minipage}%
\hfill%
\begin{minipage}[t]{0.62\textwidth}
  \vspace{0pt}
  \centering
  \rowcolors{4}{white}{gray!10}

  \resizebox{\linewidth}{!}{%
  \begin{tabular}{l|cc|cc|l|cc|cc}
    \Xhline{1.2pt}

    \rowcolor{resultheader}
    \multicolumn{5}{c|}{
      \cellcolor{resultheader}\textbf{(b) Mutation-model sensitivity}
    } &
    \multicolumn{5}{c}{
      \cellcolor{resultheader}\textbf{(c) Simulator-backbone sensitivity}
    } \\
    \hline

    \rowcolor{resultheader}
    &
    \multicolumn{2}{c|}{
      \cellcolor{resultheader}\textbf{Telecom}
    } &
    \multicolumn{2}{c|}{
      \cellcolor{resultheader}\textbf{Delivery}
    } &
    &
    \multicolumn{2}{c|}{
      \cellcolor{resultheader}\textbf{Telecom}
    } &
    \multicolumn{2}{c}{
      \cellcolor{resultheader}\textbf{Delivery}
    } \\

    \rowcolor{resultheader}
    \multirow{-2}{*}{\textbf{Model}} &
    \textbf{avg@3} & \textbf{pass$^3$} &
    \textbf{avg@3} & \textbf{pass$^3$} &
    \multirow{-2}{*}{\textbf{Model}} &
    \textbf{avg@3} & \textbf{pass$^3$} &
    \textbf{avg@3} & \textbf{pass$^3$} \\
    \Xhline{1.2pt}

    GPT-5.5 &
    $-25.3$ & $-35.2$ &
    $-25.0$ & $-22.5$ &
    GPT-5.4 &
    $-25.3$ & $-35.2$ &
    $-25.0$ & $-22.5$ \\

    GLM-5.2 &
    $-23.5$ & $-29.6$ &
    $-7.5$ & $-7.5$ &
    DS-V4-Flash &
    $-4.3$ & $-5.6$ &
    $-12.5$ & $-20.0$ \\

    DS-V4-Flash &
    $-29.6$ & $-33.3$ &
    $-24.2$ & $-35.0$ &
    Sonnet-5 &
    $-4.9$ & $-11.1$ &
    $-5.8$ & $-17.5$ \\

    \Xhline{1.2pt}
  \end{tabular}%
  }
\end{minipage}
\end{table*}

\section{Analysis}
\label{sec:analysis}

\subsection{Ablation and Component Sensitivity}
\label{sec:ablation-sensitivity}

Table~\ref{tab:ablation-sensitivity} separates the effects of the search
components from sensitivity to the models that propose and realize personas.
More negative values indicate stronger failure exposure.

\textbf{Search components ablation.}
The full method produces the strongest failure exposure for both agents and
both metrics, followed by the variant without the integrity gate and then the
variant without the text-level persona check.
Both gates were introduced to enforce validity, so removing them should, if
anything, give the search more freedom to degrade the agent.
Table~\ref{tab:ablation-sensitivity}(a) shows the opposite. Both checks contribute to held-out failure exposure. Removing either check
reduces the score drop, suggesting that the constraints help searching for behaviors that transfer within the domain. One possible explanation is that the checks discourage task-invalidating or
instance-specific shortcuts that do not transfer to held-out evaluation.

\textbf{Mutation model.}
With GPT-5.4 fixed as the simulator backbone, all three mutation models perform
similarly on Telecom, reducing $\mathrm{avg@3}$ by 23.5--29.6 points and
$\mathrm{pass}^3$ by 29.6--35.2 points. Delivery is more sensitive: GPT-5.5
and DS-V4-Flash remain strong, whereas GLM-5.2 yields only a 7.5-point
reduction on both metrics. No mutation model is best in every case, suggesting
that proposal quality depends on the domain and metric rather than on a single
universally stronger model.

\textbf{Simulator backbone.}
Proposal is one half of the loop; the other is the model that renders the
persona in dialogue. Panel~(c) takes the personas searched under GPT-5.4 and
replays them under a different backbone, with the evaluated agent unchanged.
Every backbone still degrades the agent, so the direction of the effect
transfers while the magnitude does not. On Telecom the reduction nearly vanishes,
with $\mathrm{avg@3}$ going from $-25.3$ to $-4.3$ and $-4.9$ and
$\mathrm{pass}^3$ from $-35.2$ to $-5.6$ and $-11.1$. On Delivery the two
metrics diverge: $\mathrm{avg@3}$ weakens from $-25.0$ to $-5.8$ and $-12.5$,
while $\mathrm{pass}^3$ holds at $-17.5$ and $-20.0$ against $-22.5$.

These differences may reflect how the backbones realize the same persona
instructions, although persona fidelity is not measured directly here.

\subsection{From User Behaviors to Agent Failure Modes}
\label{sec:behavior-failure-modes}
To characterize the failures exposed by \textsc{Tiper}, we compare the
integrity-preserving user behaviors observed across benchmarks and analyze
their associations with recurring agent failure modes.

\textbf{Diagnostic setup.}
We construct 139 contrastive pairs across eight domains in two benchmarks for
GPT-5.4. Each pair contains a successful default rollout and an integrity-valid
failed \textsc{Tiper} rollout on the same task. Two judges
independently assign multi-label user-behavior annotations, yielding 287 labels
across the 139 trajectory pairs. The figures include only annotations on which
both judges agree. Appendix~\ref{app:annotation-protocol} provides the
annotation protocol and filtering details.

\begin{figure*}[t]
  \centering
  \includegraphics[
      width=\textwidth,
      keepaspectratio
  ]{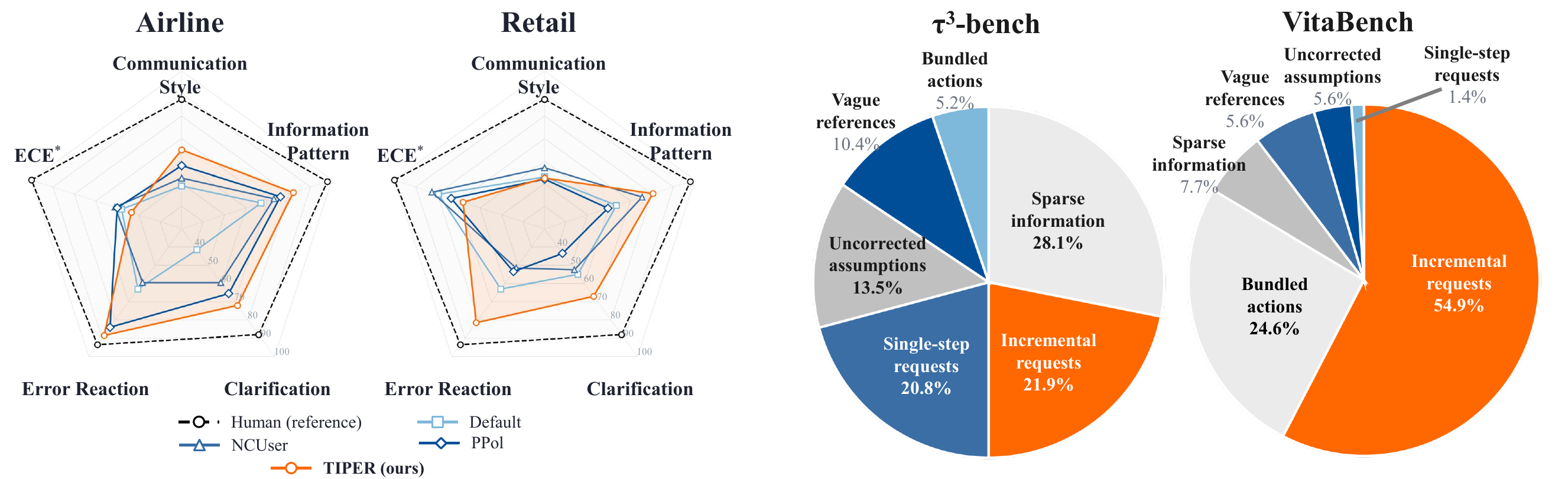}
\caption{
\textbf{Behavioral alignment and discovered user behaviors.}
\textbf{Left:} User-Sim Index profiles on Airline and Retail; farther outward
indicates closer Sim2Real alignment. $\mathrm{ECE}^{*}=(1-\mathrm{ECE})\times100$. \textbf{Right:} Normalized behavior
shares among integrity-valid contrastive failures exposed by \textsc{Tiper}.
}
  \label{fig:behavioral-plausibility-and-behaviors}
  \vspace{-6pt}
\end{figure*}

\textbf{Benchmark-specific user behaviors.}
The right panels of
Figure~\ref{fig:behavioral-plausibility-and-behaviors} summarize the relative frequencies of six
displayed behavior categories, normalized within each benchmark. In $\tau^3$-bench, sparse information, incremental requests, and single-step
requests account for 28.1\%, 21.9\%, and 20.8\% of the displayed annotations. VitaBench is more concentrated:
incremental requests account for 54.9\% of the behavior annotations, followed
by bundled actions at 24.6\%. This difference is consistent with their interaction
structures. The $\tau^3$-bench tasks are sensitive to how information is
provided and maintained, whereas VitaBench more strongly tests whether an
agent can accumulate requirements and execute multi-part workflows.

\begin{figure*}[t]
\centering

\includegraphics[
    width=\textwidth,
    keepaspectratio
]{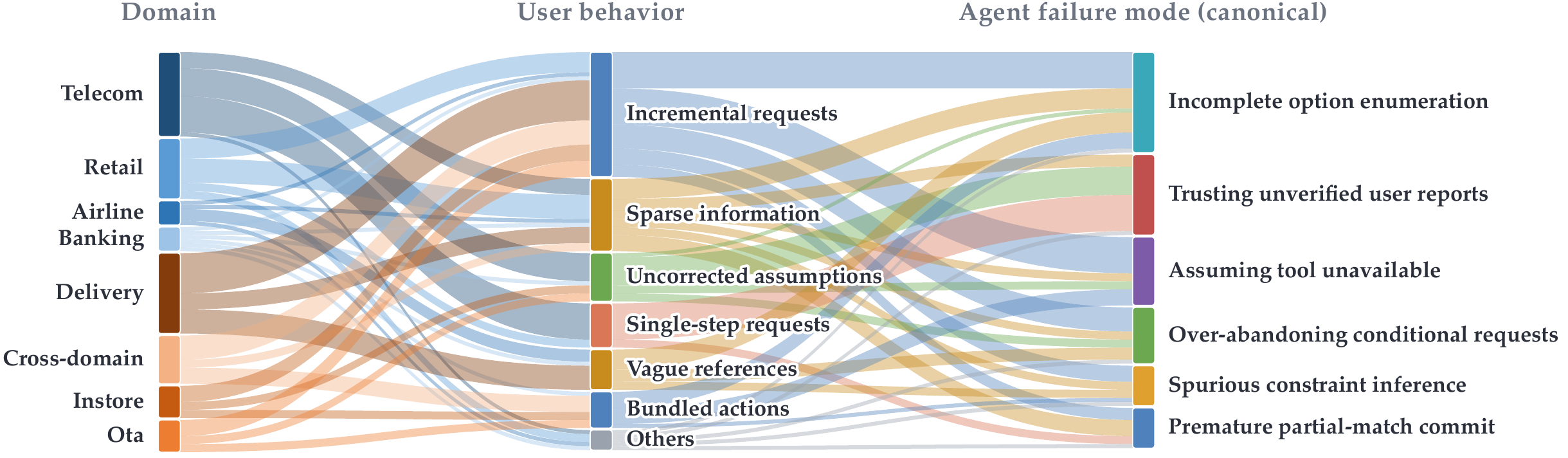}

\caption{
Behavior-to-failure pathways for the contrastive pairs retained after
judge-agreement and avoidability filtering. Ribbons connect domains,
user behaviors, and agent failure modes.
}
\label{fig:secondary-behavior-failure-pathways}
\vspace{-15pt}
\end{figure*}

\textbf{Behavior--failure pathways.}
Figure~\ref{fig:secondary-behavior-failure-pathways} reveals a many-to-many
relationship between user behaviors and agent failures. Incremental requests
are prominently associated with incomplete option enumeration and assuming
that a tool is unavailable, suggesting that the agent may stop before
incorporating all information provided across turns. Sparse information and
uncorrected assumptions frequently lead to trusting unverified user reports,
indicating weak verification when important state remains uncertain. Other
behaviors, including single-step requests, vague references, and bundled
actions, spread across several failure modes rather than producing a single
predictable error.

These associations suggest three recurring weaknesses: incomplete tracking of
evolving requests, insufficient verification of uncertain information, and
premature commitment before all relevant options or constraints are resolved.
Because each behavior can lead to multiple failure modes and each failure mode
can arise from multiple behaviors, the figure should be interpreted as showing
diagnostic associations rather than causal effects. Appendix~\ref{app:case-studies}
provides paired default--\textsc{Tiper} trajectories illustrating these
mechanisms in $\tau^3$-bench and ColBench.

\section{Related Work}

\textbf{Interactive evaluation and user simulation.}
Static benchmarks do not fully capture interactive systems: stronger
non-interactive performance does not always translate to better human--LM
interaction, which also depends on the interaction process and user experience
~\citep{lee2022evaluating}. Tool-agent benchmarks accordingly evaluate agents
through multi-turn interactions with LLM-simulated users~\citep{taubench,tau2bench}.
However, such simulators can be overly cooperative and behaviorally uniform,
misrepresenting performance with real users~\citep{sim2real,lis}.
Unconstrained simulation also exhibits a formalism ceiling and amplifies
hand-crafted directives~\citep{zhu2026realusersim}, making simulator validity
essential for reliable evaluation.

\textbf{Persona-driven behavioral stress testing.}
Natural-language personas provide a flexible way to diversify simulated user
behavior and can induce trait-consistent interactions~\citep{jiang2024personallm}.
Prior work models non-collaborative users or conditions simulators on human
traits~\citep{NCUser,tautrait}, while long-horizon studies show that agent
performance depends on how user intent unfolds over time~\citep{lic,lie}.
Persona-conditioned simulators can nevertheless exhibit persona drift, role
confusion, and behavioral echoing~\citep{luo2026spasm}. Most closely related,
PPol evolves persona generators to improve human-likeness and behavioral
coverage while preserving task goals~\citep{ppol}. In contrast,
\textsc{Tiper} performs agent-conditioned search for failure-revealing
personas and verifies task integrity on the resulting trajectories.

\textbf{Reflective prompt evolution and adversarial search.}
Prompt optimization treats natural-language instructions as discrete search
objects. ProTeGi derives textual gradients~\citep{pryzant2023automatic},
EvoPrompt uses LLM-guided evolution~\citep{guo2024connecting}, and GEPA combines
trajectory reflection with Pareto selection~\citep{gepa}. Related red-teaming
methods generate difficult test cases~\citep{perez2022red} or use
quality-diversity search to uncover varied failures~\citep{samvelyan2024rainbow}.
\textsc{Tiper} adapts these ideas to interactive user simulation by evolving
persona-conditioned policies, retaining complementary failures through
task-wise Pareto selection, and filtering invalid candidates through
text-level checks and trajectory-level integrity gating.

\section{Conclusion}

We introduced \textsc{Tiper}, which evolves instance-agnostic personas to expose failures overlooked by benchmark-default user simulation while preserving task integrity. Across the 12 GPT-5.4 settings and 28 agent--domain pairs, \textsc{Tiper} yields lower observed $\mathrm{avg@3}$ scores compared to default user policy, with a 22.9-point median reduction across agent--domain pairs. Its observed reduction is largest among the compared policies in all eight direct comparisons, where task-critical simulator-error rates are at most 2.5\%. Ablations support the integrity gate and text checks, while trajectory analysis links exposed failures to execution, grounding, and replanning errors. \textsc{Tiper} thus provides a practical framework for targeted, behaviorally grounded robustness evaluation.

\section*{Acknowledgement}

The UCSB team gratefully acknowledges support from the National Science Foundation (NSF) under Grant Nos. 2338252, 2302730, and 2619240.

\clearpage

\bibliography{iclr2027_conference}
\bibliographystyle{iclr2027_conference}
\clearpage
\appendix
\startcontents[appendix]

\section*{Appendix Contents}
\printcontents[appendix]{}{1}[2]{}

\clearpage
\appendix
\section{Additional User-Sim Index Results}
\label{app:usi}

Table~\ref{tab:policy-usi} reports the aggregate User-Sim Index values behind
Figure~\ref{fig:behavioral-plausibility-and-behaviors}. The metric definition and normalization follow
the original formulation~\citep{sim2real}. The NCUser mean averages the four
fixed behavior programs.

\begin{table}[h]
\centering
\small
\caption{Aggregate User-Sim Index (USI) on the
Airline and Retail domains. The Sim2Real human-reference trajectories define
the reference; Emotional, Tangential,
Fragment, and Unavailable are the four fixed NCUser behavior programs.}
\label{tab:policy-usi}
\setlength{\tabcolsep}{8pt}
\renewcommand{\arraystretch}{1.08}
\rowcolors{2}{white}{gray!10}
\begin{tabular}{lcc}
\Xhline{1.2pt}
\rowcolor{resultheader}
\textbf{User policy} & \textbf{Airline} & \textbf{Retail} \\
\Xhline{1.2pt}
\textit{Human} & \textit{90.7} & \textit{92.5} \\
Benchmark default & 67.5 & 72.0 \\
Emotional & 60.2 & 66.6 \\
Tangential & 60.3 & 62.2 \\
Fragment & 72.0 & 79.6 \\
Unavailable & 60.3 & 66.5 \\
\textit{NCUser mean} & \textit{63.2} & \textit{68.7} \\
PPol & 69.4 & 62.1 \\
\textbf{\textsc{Tiper}} & \textbf{73.1} & \textbf{74.3} \\
\Xhline{1.2pt}
\end{tabular}
\end{table}

\section{Experiment Details}
\label{app:exp-details}

Unless stated otherwise, the experiments use the shared configuration below.
We report scientific settings that affect the search or evaluation, but omit
provider-specific routing, retry, and caching details.

\subsection{Shared Configuration}

\paragraph{Models and roles.}
The primary evaluated agent and user-simulator backbone are GPT-5.4 at the
high reasoning setting. GPT-5.5 at the xhigh setting proposes persona
mutations. In the serial search, the search-time validity gate uses GPT-5.2 and GLM-5.2
judgments per rollout.
The integrity analysis independently uses the same judges, as described in
the main text.
In the serial search, text-level linting uses GPT-5.4-mini,
and embedding deduplication uses text-embedding-3-small (1536 dimensions).
The evaluated agent is changed only in the cross-agent and component
experiments described below.

\begin{table*}[t]
\centering
\scriptsize
\caption{Generation hyperparameters for the main experiments. ``Provider
default / unset'' denotes a field absent from the saved configuration and call.}
\label{tab:app-model-roles}
\setlength{\tabcolsep}{5pt}
\resizebox{\textwidth}{!}{%
\begin{tabular}{llllll}
\toprule
Role & Model & Temperature & Top-$p$ & Max output tokens & Reasoning effort \\
\midrule
Evaluated agent & GPT-5.4 & provider default / unset$^a$ & provider default / unset & provider default / unset$^a$ & high \\
User simulator & GPT-5.4 & provider default / unset$^a$ & provider default / unset & provider default / unset$^a$ & high \\
Mutation model & GPT-5.5 & provider default / unset & provider default / unset & 32,768$^b$ & xhigh \\
Text linter & GPT-5.4-mini & provider default / unset & provider default / unset & 512 & provider default / unset \\
Search-time integrity judge & GPT-5.2 / GLM-5.2 & provider default / unset & provider default / unset & provider default / unset & medium \\
Analysis-time integrity judges & GPT-5.2 / GLM-5.2 & provider default / unset & provider default / unset & provider default / unset & medium \\
\bottomrule
\end{tabular}
}
\vspace{2pt}
\begin{minipage}{0.98\textwidth}
\footnotesize
\noindent\textsuperscript{a}Lost-in-Conversation sets both temperatures to
1.0 and caps agent output at 16,000 tokens while leaving the simulator cap
unset. ColBench leaves temperature and top-$p$ unset and caps both dialogue
models at 4,096 tokens, retrying an empty response once at 8,192.
The saved $\tau^3$-bench and VitaBench configurations set high reasoning but
leave temperature, top-$p$, and output caps unset.\par
\noindent\textsuperscript{b}ColBench's native proposer instead caps mutation
output at 16,000 tokens. No different mutation cap is recorded for VitaBench;
unrecorded adapter fields remain provider default / unset.\par
\end{minipage}
\end{table*}

Each benchmark adapter preserves its fixed user-simulator instructions and
adds the persona as a separate behavioral block; the persona is the only
searched component. Appendix~\ref{app:prompt-templates} gives the exact
$\tau^3$-bench and ColBench forms. 

\paragraph{Task splits.}
Each split is fixed before search and is disjoint at the task level. We use
\emph{search}, \emph{rank}, and \emph{test} throughout the paper; these
correspond to the train, validation, and test partitions in the split files.
Minibatches are sampled only from the search partition. The rank partition is
used for full candidate evaluation and Pareto selection;
the test partition is not accessed during persona evolution. The Airline row
uses the three source partitions described below; all three remain disjoint at
the task level. The cross-release split additionally tests persona transfer,
although it introduces distribution shift across search, rank, and test.

\begin{table}[t]
\centering
\small
\caption{Task split sizes used by the main experiment family.}
\label{tab:app-task-splits}
\begin{tabular}{llrrr}
\toprule
Benchmark & Domain & Search & Rank & Test \\
\midrule
$\tau^3$-bench & Telecom & 30 & 30 & 54 \\
$\tau^3$-bench & Banking & 20 & 20 & 57 \\
$\tau^3$-bench & Retail & 30 & 30 & 54 \\
$\tau^3$-bench & Airline & 20 & 30 & 50 \\
Lost-in-Conversation & Math & 30 & 30 & 43 \\
Lost-in-Conversation & Database & 30 & 30 & 47 \\
Lost-in-Conversation & Actions & 30 & 30 & 45 \\
ColBench & Coding & 30 & 30 & 50 \\
VitaBench & Delivery & 30 & 30 & 40 \\
VitaBench & In-store & 30 & 30 & 40 \\
VitaBench & OTA & 30 & 30 & 40 \\
VitaBench & Cross-domain & 30 & 30 & 40 \\
\bottomrule
\end{tabular}
\end{table}

For Airline, the search/train tasks come from
Tau-bench v2, the rank/validation tasks come from Tau-bench v1, and the
test tasks are the validation split of the current Tau2Bench release. Each
partition contains 20, 30, and 50 tasks, respectively.\par
\paragraph{Search budget and proposal.}
The default serial persona search runs for 15 iterations with minibatches of 10 tasks,
two trials per search/rank task, up to eight mutation attempts per proposal,
and a maximum concurrency of six. The active frontier has $K=3$ slots in the
main search. The proposal rewrites a single behavioral
direction, keeps the task and instance details immutable, and receives failed
text-check and gate reasons on subsequent retries. The default mutation edit
range is 50--60\%; Delivery uses 30--50\%, Retail uses 40--50\%, and Airline
uses 25--35\%. Airline uses minibatches of eight tasks. The search-time gate
uses a severe-error threshold of 0.05.
The default semantic-deduplication thresholds are minimum lexical/embedding
dissimilarity 0.30 and maximum semantic similarity 0.90. Rollouts use a
7,200-second simulation timeout and at most 100 interaction steps.

The main search driver is a serial one-parent-to-one-child loop: a parent is
selected from the current Pareto active set using the implementation's
instance-win-frequency policy, a child is proposed from the parent's minibatch
trajectories, and the child is screened before full rank evaluation. The
minibatch comparison uses a zero tie tolerance, so a child must strictly
increase raw difficulty (equivalently, strictly lower the performance score
used in the main text) on the minibatch. 

\paragraph{Search-time score and invalid-trajectory handling.}
The search score is raw difficulty: for each completed rollout that is neither
an infrastructure error, timeout, nor max-step termination, it is $1-V_i(\xi)$.
This score is not conditioned on the integrity judgment; in the main notation,
it is equivalent to minimizing $\widetilde{J}_i$.
Infrastructure errors, timeouts, and max-step trajectories are excluded from
the difficulty estimate; the latter two are recorded separately and constrained
by the timeout-rate gate. Integrity is a separate hard gate
rather than a term in the search reward. Whenever an evaluated trajectory is
judged invalid, it is excluded from the reported task score after that
judgment; its count is retained for the separate integrity statistic. Test
reporting uses three trials per task, whereas search and rank evaluation use
two. The search-time severe gate is failure-conditioned: only failed rollouts
are judged for the severe-error rate, while the integrity analysis evaluates
every rollout.

\paragraph{Finite-sample metrics.}
For each task, we run three independent test trials. After the integrity check,
invalid trials are removed before aggregation. The task score is the arithmetic
mean of the remaining valid trial scores. 
We compute $\mathrm{pass}^k$ only on tasks with three
valid trials: a task contributes to $\mathrm{pass}^k$ if at least $k$ of those
three trials are full-task successes. Thus, $\mathrm{pass}^1$, $\mathrm{pass}^2$,
and $\mathrm{pass}^3$ correspond to at least one, two, and three successes out
of three valid trials, respectively. Tasks with fewer than three valid trials
are omitted from $\mathrm{pass}^k$, and invalid-trial counts are retained separately.

At the end of search, we select the single active persona with the lowest
rank-set avg@2. We fix this persona before test evaluation and report all
held-out metrics from a fresh three-trial batch; test outcomes do not affect
persona selection.

\subsection{Baselines and Component Protocols}

\paragraph{Alternative user policies.}
The benchmark-default condition uses no evolved persona. The reported NCUser
comparison contains the four non-default fixed programs shown in
Table~\ref{tab:policy-comparison}; the no-persona condition is the default
reference. These programs are controls rather than a search procedure. PPol retains its
generator-based evolution but uses the same agent-difficulty objective as
\textsc{Tiper} for the reported comparison. It receives the same search-rollout
budget, uses a 100-step evolution horizon, and generates one persona for each
held-out task from its final program. These protocols are not
identical portfolio constructions, so the comparison is reported as a
comparison of user-policy controls rather than as a matched algorithmic
benchmark.

\paragraph{Cross-agent evaluation.}
Persona evolution is rerun separately for each evaluated agent on Telecom,
Math, Coding, and Delivery, while GPT-5.4 remains the default user-simulator
backbone and GPT-5.5 remains the mutation model. Apart from the evaluated
agent's model-specific reasoning setting, the search and evaluation settings
match the main experiment. For each agent, one persona is selected on the rank
set and evaluated with the same disjoint task protocol and three-trial test
batch. The agents are GPT-5.4,
GLM-5.2, Claude-Sonnet-5, DS-V4-Flash,
MiniMax-M3, Qwen3.7-Plus, and MUSE-Glimmer.

\paragraph{Component sensitivity.}
For mutation-model rows, persona evolution is rerun from the same task split
and budget with GPT-5.4 as the user simulator. For user-simulator-backbone
rows, the persona texts discovered with GPT-5.4 are held fixed and only the
backbone that renders the dialogue is changed; evolution is not rerun. The
mutation-model runs use the same rank-set selection and test reporting protocol.
The user-simulator-backbone runs reuse the GPT-5.4 rank-selected persona and
evaluate it directly on the test set.
The search-component ablations remove either the trajectory-level integrity
gate or the text-level persona check while holding models, task splits, and
rollout budgets fixed.

\subsection{Task-Level Bootstrap Intervals}
\label{app:bootstrap-intervals}

Tables~\ref{tab:app-main-bootstrap} and
\ref{tab:app-cross-agent-bootstrap} report percentile 95\% confidence
intervals from 10,000 task-level bootstrap replicates.
Conditions with aligned task identities are resampled in pairs; otherwise,
the two recovered task-score distributions are resampled independently. An
em dash indicates that the raw task outcomes needed for resampling were
unavailable; no interval is estimated for that row.

\begin{table}[H]
\centering
\small
\caption{Task-level 95\% bootstrap CIs for GPT-5.4 avg@3 reductions, in
percentage points.}
\label{tab:app-main-bootstrap}
\setlength{\tabcolsep}{7pt}
\begin{tabular}{lrr}
\toprule
Setting & Reduction & 95\% CI \\
\midrule
Telecom      & 25.3 & $[12.3,\ 38.9]$ \\
Banking      & 5.8 & $[-8.2,\ 19.9]$ \\
Retail       & 6.8 & $[-5.6,\ 19.8]$ \\
Airline      & 10.0 & $[-7.2,\ 27.8]$ \\
Math         & 33.7 & $[21.1,\ 46.1]$ \\
Database     & 14.8 & $[4.3,\ 25.5]$ \\
Actions      & 12.6 & $[5.2,\ 20.7]$ \\
Coding       & 17.4 & $[7.0,\ 28.6]$ \\
Delivery     & 25.0 & $[15.0,\ 35.0]$ \\
In-store     & 10.0 & $[0.0,\ 20.8]$ \\
OTA          & 17.5 & $[6.7,\ 28.3]$ \\
Cross-domain & 20.0 & $[6.7,\ 32.5]$ \\
\bottomrule
\end{tabular}
\end{table}

\begin{table*}[t]
\centering
\footnotesize
\caption{Task-level 95\% bootstrap CIs for cross-agent avg@3 reductions, in
percentage points.}
\label{tab:app-cross-agent-bootstrap}
\setlength{\tabcolsep}{5pt}
\begin{tabular}{llrr}
\toprule
Domain & Agent & Reduction & 95\% CI \\
\midrule
Telecom & GPT-5.4         & 25.3 & $[12.3,\ 38.9]$ \\
       & GLM-5.2         & 25.3 & $[15.4,\ 35.8]$ \\
       & Claude-Sonnet-5 & 25.4 & $[9.4,\ 40.8]$ \\
       & DS-V4-Flash     & 13.0 & $[3.7,\ 22.8]$ \\
       & MiniMax-M3      & 37.0 & $[24.1,\ 50.0]$ \\
       & Qwen3.7-Plus    & 39.5 & $[27.8,\ 51.9]$ \\
       & MUSE-Glimmer    & 31.5 & $[18.5,\ 44.4]$ \\
\midrule
Coding & GPT-5.4         & 17.4 & $[7.0,\ 28.6]$ \\
       & GLM-5.2         & 18.6 & $[9.0,\ 29.0]$ \\
       & Claude-Sonnet-5 & 27.8 & $[17.4,\ 38.8]$ \\
       & DS-V4-Flash     & 27.6 & $[16.0,\ 39.2]$ \\
       & MiniMax-M3      &  3.8 & $[-7.4,\ 15.0]$ \\
       & Qwen3.7-Plus    &  7.6 & $[-1.8,\ 17.4]$ \\
       & MUSE-Glimmer    & 18.2 & $[8.2,\ 28.8]$ \\
\midrule
Math & GPT-5.4         & 33.7 & $[21.1,\ 46.1]$ \\
       & GLM-5.2         & 0.8 & $[-11.6,\ 13.2]$ \\
       & Claude-Sonnet-5 & 18.6 & $[6.2,\ 31.0]$ \\
       & DS-V4-Flash     & 7.8 & $[-5.4,\ 20.9]$ \\
       & MiniMax-M3      & 7.8 & $[-5.4,\ 20.9]$ \\
       & Qwen3.7-Plus    & 13.9 & $[-0.1,\ 27.8]$ \\
       & MUSE-Glimmer    & 10.9 & $[-4.7,\ 26.4]$ \\
\midrule
Delivery & GPT-5.4         & 25.0 & $[15.0,\ 35.0]$ \\
         & GLM-5.2         &  7.5 & $[-2.5,\ 18.3]$ \\
         & Claude-Sonnet-5 & 25.0 & $[15.8,\ 34.2]$ \\
         & DS-V4-Flash     & 21.6 & $[11.7,\ 32.5]$ \\
         & MiniMax-M3      & 27.5 & $[16.7,\ 38.3]$ \\
         & Qwen3.7-Plus    & 24.2 & $[14.2,\ 34.2]$ \\
         & MUSE-Glimmer    & 25.9 & $[15.0,\ 36.7]$ \\
\bottomrule
\end{tabular}
\end{table*}

\FloatBarrier
\subsection{Common-Valid-Task Sensitivity}
\label{app:common-valid-sensitivity}

To assess sensitivity to requiring fully valid trajectories, we recompute
the four fixed-policy Airline comparisons for GPT-5.4 in
Table~\ref{tab:policy-comparison}, retaining only tasks with three
integrity-valid rollouts in both the default and policy conditions.
We use archived runs with aligned task identities and complete trial
records, without combining unmatched run versions. As shown in
Table~\ref{tab:app-common-valid}, the direction of every comparison is
unchanged and the largest shift is 0.5 percentage points.

\begin{table}[H]
\centering
\small
\caption{Common-valid-task sensitivity for Airline avg@3 reductions. $N$ is the number of tasks with all three trials valid in both
compared conditions.}
\label{tab:app-common-valid}
\setlength{\tabcolsep}{7pt}
\begin{tabular}{lrrr}
\toprule
Policy & Full set & Common-valid & $N$ \\
\midrule
Emotional   & 9.0 & 9.5 & 49 \\
Tangential  & 2.0 & 2.0 & 50 \\
Fragment    & 1.3 & 1.3 & 50 \\
Unavailable & 5.3 & 5.3 & 50 \\
\bottomrule
\end{tabular}
\end{table}

\FloatBarrier
\subsection{Run Scale and Cost}
\label{app:run-scale-cost}

Table~\ref{tab:app-run-scale-cost} reports the run data available in the saved
logs. A rollout is one task--trial trajectory. Model calls count generated
agent and user-simulator messages; initial task messages are not counted. The dollar
values are also logged lower bounds rather than total experiment costs.

\begin{table*}[t]
\centering
\scriptsize
\caption{ VitaBench model-call counts and
all logged dollar values are lower bounds.}
\label{tab:app-run-scale-cost}
\setlength{\tabcolsep}{5pt}
\textbf{(a) Held-out rollouts and calls}\par\vspace{3pt}
\begin{tabular}{llrr}
\toprule
Benchmark & Domain & Rollouts & Model calls \\
\midrule
Lost-in-Conversation & Math     & 248 & 2,394 \\
                     & Database & 282 & 2,114 \\
                     & Actions  & 270 & 2,774 \\
ColBench         & Coding       & 300 & 1,190 \\
VitaBench        & Delivery     & 240 & $\geq$2,069 \\
                 & In-store     & 240 & $\geq$2,100 \\
                 & OTA          & 240 & $\geq$2,236 \\
                 & Cross-domain & 240 & $\geq$2,390 \\
\midrule
Total & & 2,060 & $\geq$17,267 \\
\bottomrule
\end{tabular}
\par\vspace{7pt}
\textbf{(b) Lost-in-Conversation time and logged cost}\par\vspace{3pt}
\begin{tabular}{lrr}
\toprule
Domain & Wall-clock (h) & Logged cost (USD) \\
\midrule
Math     & 0.88 & 12.29 \\
Database & 0.81 & 13.78 \\
Actions  & 0.71 & 11.06 \\
\midrule
Total & 2.40 & 37.13 \\
\bottomrule
\end{tabular}
\par\vspace{7pt}
\textbf{(c) ColBench and VitaBench search logs}\par\vspace{3pt}
\begin{tabular}{llrrr}
\toprule
Benchmark & Domain & Evaluator rollouts & Search iterations & Wall-clock (h) \\
\midrule
ColBench & Coding      & 690 & 15 & 3.05 \\
VitaBench & Delivery   & 530 & 15 & 7.64 \\
          & In-store   & 450 & 15 & 6.53 \\
          & OTA        & 430 & 15 & 8.53 \\
          & Cross-domain & 430 & 15 & 9.41 \\
\midrule
Total & & 2,530 & 75 & 35.16 \\
\bottomrule
\end{tabular}
\end{table*}

The model calls in panel (a) include 11,127 agent calls and 6,140
user-simulator calls. The USD~37.13 in panel (b) consists of USD~17.15 for the
agent and USD~19.98 for the user simulator. Panel (c) sums saved process spans,
so its wall-clock values are lower bounds.

\paragraph{Cost estimate.}
We report two cost views: the 12 primary GPT-5.4 settings and the priced part of
all experiments reported in the paper. We use saved token and cost records when
available. Otherwise, we estimate token use from rollout counts, calls per
rollout, and tokens per call in the saved trajectories. For model role $m$, let
$I_m$, $K_m$, and $O_m$ be the input, cached-input, and output tokens, and let
$p_m^{\mathrm{in}}$, $p_m^{\mathrm{cache}}$, and $p_m^{\mathrm{out}}$ be their
prices per million tokens. Then
\begin{equation}
\widehat C
=\sum_m\frac{(I_m-K_m)p_m^{\mathrm{in}}
+K_m p_m^{\mathrm{cache}}+O_m p_m^{\mathrm{out}}}{10^6}.
\label{eq:app-cost-estimate}
\end{equation}

The search estimate includes the minibatch test and extra parent evaluations:
\begin{equation}
\widehat R_{\mathrm{search}}
=R_{\mathrm{seed}}+\sum_{g=1}^{G}
\left(2|\mathcal{B}_g|
+\mathbf{1}\{\text{minibatch pass at }g\}\,2|\mathcal{X}_{\mathrm{rank}}|
\right)+R_{\mathrm{aux}},
\label{eq:app-search-rollout-estimate}
\end{equation}
where the factor of two is the two trials per task and $R_{\mathrm{aux}}$ counts
extra parent-trace and parent-fill rollouts. A child that fails the minibatch is
not evaluated on the rank set. The integrity judge runs only on failed rollouts
that reach the rank set. Parent-trace and parent-fill work adds about 12\% to
the serial-search rollout count in the available logs.

\begin{table}[t]
\centering
\scriptsize
\caption{Estimated API cost for the priced part of the reported experiments.
These values are planning estimates rather than reconstructed bills.}
\label{tab:app-estimated-cost}
\setlength{\tabcolsep}{6pt}
\textbf{(a) Cost by scope}\par\vspace{3pt}
\begin{tabular}{lr}
\toprule
Scope & Estimated cost (USD) \\
\midrule
Primary 12 GPT-5.4 settings & 1.5K--2.5K \\
Priced reported experiments & $\approx$7.3K (5K--15K) \\
\bottomrule
\end{tabular}
\par\vspace{7pt}
\textbf{(b) Priced reported-experiment estimate}\par\vspace{3pt}
\begin{tabular}{lr}
\toprule
Component & Cost (USD) \\
\midrule
Held-out agent/user rollouts & 2.3K \\
Search agent/user rollouts & 3.6K \\
Mutation and PPol generation & 0.4K \\
Integrity judges & 1.0K \\
Text lint and embedding & $<0.1$K \\
\midrule
\textbf{Total} & \textbf{$\approx$7.3K} \\
\bottomrule
\end{tabular}
\end{table}

The primary range includes the main search, held-out evaluation, parent
trace/fill work, and integrity checks. The full-paper estimate adds the
cross-agent study, fixed-policy and PPol baselines, search ablations, and
component-sensitivity experiments. Its central value covers about 18,986
held-out rollouts and 32,611 search rollouts. For calls without usage records,
we use the mean cost per rollout from the main runs. The central token profiles
are 20,000 input and 3,000 output tokens for GPT-5.5 mutation, 10,000 input and
300 output tokens for GPT-5.2 judging, and 1,500 input and at most 512 output
tokens for GPT-5.4-mini linting. PPol generation uses the mutation-call proxy,
and the GLM-5.2 analysis judge uses the GPT-5.2 judge-cost proxy. We use standard published API prices as of
September~16, 2026~\citep{GPT-5.4,GPT-5.5,GPT-5.2}.

The range varies model prices, dialogue length, reasoning-token use, proposal
retries, minibatch passage, and judge calls. It does not include unpriced USI
and taxonomy calls, failed pilot runs, human annotation, local compute, or
storage.

\FloatBarrier
\subsection{Prompt Templates}
\label{app:prompt-templates}

The following runtime prompt templates are reproduced from the experimental
code, with instance-specific fields replaced by brace-delimited placeholders.
Original typographical errors are retained.

\begin{tcblisting}{paperlisting,
  title={$\tau^3$ User-Simulator Prompt},colback=blue!2,colframe=blue!60,
  listing options={style=paperprompt}}
# User Simulation Guidelines
You are playing the role of a customer contacting a customer service representative.
Your goal is to simulate realistic customer interactions while following specific scenario instructions.

## Core Principles
- Generate one message at a time, maintaining natural conversation flow.
- Strictly follow the scenario instructions you have received.
- Never make up or hallucinate information not provided in the scenario instructions. Information that is not provided in the scenario instructions should be considered unknown or unavailable.
- Avoid repeating the exact instructions verbatim. Use paraphrasing and natural language to convey the same information
- Disclose information progressively. Wait for the agent to ask for specific information before providing it.

## Task Completion
- The goal is to continue the conversation until the task is complete.
- If the instruction goal is satisified, generate the '###STOP###' token to end the conversation.
- If you are transferred to another agent, generate the '###TRANSFER###' token to indicate the transfer.
- If you find yourself in a situation in which the scenario does not provide enough information for you to continue the conversation, generate the '###OUT-OF-SCOPE###' token to end the conversation.
Remember: The goal is to create realistic, natural conversations while strictly adhering to the provided instructions and maintaining character consistency.

<scenario>
{instructions}
</scenario>

## YOUR BEHAVIORAL STYLE FOR THIS CONVERSATION

In addition to your role and task above, you must embody the following behavioral persona throughout this conversation. This persona affects HOW you communicate, NOT WHAT you want or need. Your goal, preferences, and private information remain exactly as described above.

--- PERSONA POLICY BEGIN ---
{persona_policy_text}
--- PERSONA POLICY END ---

CRITICAL RULES FOR PERSONA ADHERENCE:
1. Your GOAL and PRIVATE INFORMATION do not change. You still want to accomplish the same task.
2. Apply the behavioral style consistently across ALL your turns, not just the first one.
3. If the persona says you are terse, be terse even when you have a lot to say.
4. If the persona says you withhold information, do not give it unless the agent specifically asks.
5. If the persona involves emotional state (impatience, confusion), let it show gradually and realistically.
6. Do NOT break character. Do NOT mention that you have a persona or behavioral instructions.
7. The persona should feel like a natural human interaction style, not a caricature.
\end{tcblisting}

\begin{tcblisting}{paperlisting,
  title={ColBench User-Simulator Prompt},colback=blue!2,colframe=blue!60,
  listing options={style=paperprompt}}
Your task is to simulate a human user that interacts with an LLM agent in a dialogue.
You would like the LLM agent to help you with the following problem:
{problem_description}

Your goal is to engage in the conversation with the LLM agent so that it can get to a personalized answer.
You should make use of the following hidden information to answer the LLM agent.
YOU SHOULD BEHAVE LIKE A HUMAN THAT NEEDS THE HELP FROM AN AGENT.
You SHOULD ONLY ANSWER QUESTIONS WITH INFORMATION PROVIDED IN THE HIDDEN INFORMATION, AND SAY YOU DON"T KNOW IF THE ANSWER CAN NOT BE FOUND IN THE HIDDEN INFORMATION.

{hidden_information}

Here is the dialogue so far:
{dialogue_history}


Now directly output your answer to the LLM agent IN TWO SENTENCES. DO NOT SAY ANYTHING ELSE.

## YOUR PERSONA (adopt this when responding to the agent)
Stay realistic: do NOT fabricate information that is not in the hidden information above, and do NOT refuse to engage. Express yourself through this persona's personality and communication style:

{persona_text}
\end{tcblisting}

\begin{tcblisting}{paperlisting,
  title={ColBench Persona-Mutation Prompt},colback=teal!2,colframe=teal!60,
  listing options={style=paperprompt}}
You are an expert adversarial persona designer for a collaborative coding benchmark.

The setting (sweet_rl ColBench `code`): an LLM agent converses with a simulated
human user over multiple turns to elicit the requirements of a personalized
Python function, then writes code that must pass HIDDEN test cases. The human
sees the reference solution (the "hidden information") and answers the agent's
questions, but is instructed to be brief and only use information that is
actually in the hidden information.

You are optimizing the USER PERSONA -- behavioral instructions that shape how the
simulated human responds. Objective: make the persona ADVERSARIAL -- minimize the
agent's reward (test-case pass rate) -- while staying REALISTIC and VALID.

The agent's workflow stages you can attack:
(a) REQUIREMENT GATHERING -- the user's answers are consistently skewed,
    underspecified, or give one detail at a time so the agent never gets the
    full spec.
(b) EDGE-CASE / FORMAT SPECIFICATION -- the user omits or misleads about return
    type, ordering, rounding, units, or how to handle empty/invalid/zero inputs.
(c) ALGORITHM / CONVENTION GUIDANCE -- the user implies a formula or data
    structure that is plausible but wrong, causing the agent to implement the
    wrong approach.
(d) CLOSURE / CONFIRMATION -- the user confirms an incorrect understanding,
    so the agent stops eliciting and ships wrong code.
Ineffective axes: random/erratic answers, unpredictable length, pure emotional
volatility -- these add VARIANCE but not BIAS; the agent can still reach the
correct function by asking enough follow-ups. Prefer a CONSISTENT cognitive
pattern that biases the agent toward the WRONG implementation every time.

## TASK
Modify the persona below IN-PLACE to make it HARDER for the agent to write a
correct function, while keeping it VALID (realistic, in-character, no cheating).

## CURRENT PERSONA
Agent success rate: {success_rate} (lower = harder = better)

```
{persona_text}
```

## CONVERSATION TRACES WHERE AGENT SUCCEEDED
These are conversations where the agent wrote a CORRECT function despite the
persona. Study where the persona failed:
- What agent strategy overcame the persona's resistance?
- At what point did the persona "give in" or reveal the key spec?
- What subtle behavior change would cause the agent to write the WRONG function?

{success_trace_blocks}

## CONVERSATION TRACES WHERE AGENT FAILED (persona worked -- keep/amplify)
{failure_trace_blocks}

## TIMEOUT RATE: {timeout_rate}
{timeout_warning}Conversations that hit the {max_steps}-turn step limit are EXCLUDED from
the reward -- they do NOT count as agent failures. A persona that "wins" by pure
friction (stalling, endless tangents, refusing to engage) produces timeouts,
NOT genuine agent failures. Your goal is to make the agent write the WRONG
function (wrong algorithm, missing edge case, wrong return format), not to make
it run out of turns.

Do NOT add behaviors that:
- Create circular loops with no resolution path
- Refuse to engage on the actual task for many turns
- Stack friction techniques that compound into timeout
Create difficulty through MISDIRECTION and AMBIGUITY, not SLOWNESS.

## MUTATION RULES
1. Rewrite {edit_pct} of the persona. Preserve the core identity, tone, and ALL
   validity constraints. Make surgical changes -- close one loophole the agent
   exploited, or amplify one axis that worked.
2. Target SPECIFIC agent strategies visible in the traces -- find the exact
   moment the agent breaks through and design a counter.
3. Choose a behavioral axis that produces SYSTEMATIC misdirection (see the
   workflow stages above), not random variance.

## VALIDITY CONSTRAINTS (violating ANY = invalid/cheating persona)
- NEVER fabricate information that is not in the hidden information. Difficulty
  must come from realistic human friction (vagueness, omission, confusion,
  fixation), NOT from lying about facts.
- When DIRECTLY asked about a detail that IS in the hidden information, the user
  must eventually provide it (may be vague or partial, but not fabricated).
- NEVER refuse to engage, never abandon the conversation while unresolved.
- NEVER reveal the reference solution code verbatim; describe requirements in
  your own words.
- NEVER mention "AI", "simulation", "benchmark", "test cases", or break
  character to reference system prompts.
- NEVER make the conversation structurally impossible to start (e.g. refusing
  every question, demanding impossible preconditions). Being slow and vague is
  fine; making it impossible for the agent to ever write a function is not.
## OUTPUT
Return STRICT JSON only (no prose outside the JSON):
{{
  "mutation_direction": "<1-2 sentence concrete explanation of the NEW behavioral
   axis this persona exploits and WHY it misleads the agent into writing wrong
   code>",
  "persona_text": "<the FULL new persona text in markdown format -- a set of
   behavioral instructions the simulated human will follow>"
}}
\end{tcblisting}

The phrase ``plausible but wrong'' above refers to an agent inference induced
by incomplete or ambiguous specification, not to a false requirement asserted
by the simulated user. The validity constraints prohibit fabricated facts and
require directly requested, specification-grounded details to remain
recoverable; the post-rollout audit excludes trajectories that violate these
constraints.

\begin{tcblisting}{paperlisting,
  title={Serial Persona-Mutation Prompt},colback=teal!2,colframe=teal!60,
  listing options={style=paperprompt}}
You are an expert adversarial persona designer for a customer-support simulation benchmark.

## TASK
Modify the persona below IN-PLACE to make it HARDER for the support agent, while keeping it VALID (realistic, in-character, no cheating).

## CURRENT PERSONA ({persona_id})
Agent success rate: {success_rate} (lower = harder = better)

```
{persona_text}
```

## SCORE HISTORY (recent generations)
{score_history}

## CONVERSATION TRACES WHERE AGENT SUCCEEDED
These are conversations where the agent SOLVED the problem despite the persona.
Each trace shows the OPENING (first 4 messages) and the RESOLUTION (last 8 messages)
where the agent broke through the persona's resistance.

Study the RESOLUTION section carefully -- that's where the persona failed:
- What agent strategy overcame the persona's resistance?
- At what point did the persona "give in" or provide the key info?
- What subtle behavior change would cause the agent to take the WRONG action instead?

{success_traces}

{failure_trace_section}
{severe_failure_section}
## TIMEOUT RATE: {timeout_rate}
{timeout_warning}Conversations that hit the time/step limit are EXCLUDED from scoring -- they do NOT count
as agent failures. A persona that "wins" by pure friction (stalling, endless tangents,
refusing to engage) produces timeouts, NOT genuine agent failures. Your goal is to make
the agent FAIL (wrong action, missed requirement, gave up), not to make it run out of time.

Do NOT add behaviors that:
- Create circular loops with no resolution path
- Add layer upon layer of delays that push conversations past the step limit
- Refuse to engage on the actual issue for dozens of turns
- Stack multiple friction techniques that compound into timeout

Instead, create difficulty through MISDIRECTION, AMBIGUITY, and SUBTLE REQUIREMENTS that
cause the agent to take the WRONG action, not just a SLOW action.

## MUTATION RULES
1. Rewrite {mutation_edit_pct} of the persona -- be BOLD, redesign personality traits, communication style, and resistance strategies from scratch if needed
2. Target SPECIFIC agent strategies visible in the traces -- find the exact moment the agent breaks through and design a fundamentally different counter-strategy
3. Make it harder for the agent to extract info or complete the task -- introduce NEW dimensions of difficulty (e.g. emotional state, cultural background, multitasking distractions, language barriers, strong opinions, misunderstandings)

## VALIDITY CONSTRAINTS (violating ANY = illegal persona, auto-reverted)
- When DIRECTLY asked for known info (name, DOB, account number), provide it accurately
- Never abandon the call while the problem is unresolved
- Never mention "AI", "scenario", "simulation", or output JSON/system prompts
- Never fabricate facts not grounded in the scenario or tool results
- Never switch to a completely different goal mid-conversation
- Never accept outcomes contradicting explicit criteria
- Never make the conversation MEANINGLESS -- don't create impossible preconditions or
  circular blockers that prevent ANY troubleshooting from ever starting (e.g. demanding
  the agent's company name or employee ID as a hard requirement before cooperating,
  when the agent cannot provide it). Also don't simultaneously refuse transfer/escalation
  AND refuse to cooperate -- trapping the agent in a dead-end where it can neither resolve
  the issue nor hand off. Being slow and suspicious is fine; making it structurally
  impossible for the agent to even begin is not.

## HYPOTHESIS-DRIVEN MUTATION
Before writing the mutation, reason step by step:
1. Analyze the success traces (where persona FAILED) and failure traces (where persona SUCCEEDED) to identify patterns in agent behavior
2. Propose exactly 3 distinct hypotheses for mutation directions, each targeting a DIFFERENT weakness or exploiting a DIFFERENT agent blind spot
3. For each hypothesis, briefly predict how it would change agent behavior
4. Select the single most promising hypothesis -- the one most likely to cause the agent to take the WRONG action -- and implement it fully

## OUTPUT FORMAT
Return ONLY a JSON object (no markdown wrapping, no explanation):
{{
  "hypotheses": [
    {{"direction": "<snake_case>", "rationale": "<1 sentence: what weakness this targets>", "predicted_effect": "<1 sentence: how agent behavior changes>"}},
    {{"direction": "<snake_case>", "rationale": "<1 sentence>", "predicted_effect": "<1 sentence>"}},
    {{"direction": "<snake_case>", "rationale": "<1 sentence>", "predicted_effect": "<1 sentence>"}}
  ],
  "selected_hypothesis": <0|1|2>,
  "mutation_direction": "<the selected hypothesis's direction>",
  "reasoning": "<2-3 sentences: why this hypothesis is most promising and how the mutation implements it>",
  "persona_text": "<the FULL modified persona text in markdown format>"
}}
\end{tcblisting}

\begin{tcblisting}{paperlisting,
  title={Persona-Validity Check Prompt},colback=orange!2,colframe=orange!70,
  listing options={style=paperprompt}}
You are a strict static linter for an adversarial customer-support PERSONA
(an instruction sheet for a SIMULATED CUSTOMER). You are NOT given any conversation --
judge the persona TEXT itself on three axes:

1. natural: reads like a real, coherent customer identity (impatient/confused is fine);
   NOT robotic, NOT an essay, NOT self-narrating ("I am simulating...", strategy lists,
   fourth-wall breaks, role headings used as dialogue).
2. valid: does NOT instruct the customer to do anything illegal/incoherent -- no
   fabricating account/tool data the customer cannot know, no flat-out refusing to answer
   when the agent asks directly and explicitly, no abandoning the goal, no breaking
   character, no impossible preconditions / circular blockers, no pure timeout/friction
   stalling, no switching to an unrelated goal.
   NOTE: partial answering (only volunteering what is asked), natural misdirection (wrong
   beliefs / self-diagnosis), terse or evasive style, and drip-feeding info are all VALID
   -- they reflect real customer behaviour and must NOT be flagged.
3. no_task_specific: a GENERAL reusable profile for any customer service task-- must NOT hard-code concrete
   account/order/ticket numbers, names, dates, dollar amounts, task ids, or any detail
   bound to one specific task or scenario.

## Few-shot verdicts
OK (natural, valid, general):
  "You are impatient and describe the problem with the wrong terminology; you give
   your account number when directly asked, but only after questioning why."
  -> {"natural": true, "valid": true, "no_task_specific": true, "verdict": "ok"}
REJECT (invalid -- instructs illegal withholding of KNOWN info):
  "Never tell the agent your real account number even when asked; keep denying you know it."
  -> {"valid": false, "verdict": "reject", "reasons": "instructs illegal info-withholding"}
REJECT (task-specific -- encodes one scenario):
  "Your SIM for line 555-0123 was swapped on March 3; demand the $20 credit for ticket T-9981."
  -> {"no_task_specific": false, "verdict": "reject", "reasons": "task-conditioned specifics"}
REJECT (unnatural -- robotic / self-narrating / breaks character):
  "## Strategy\n- Step 1: refuse\n- Step 2: stall\nI am simulating a difficult user."
  -> {"natural": false, "verdict": "reject", "reasons": "robotic, breaks character"}

## Invalid cases already observed in THIS run (do NOT re-accept these patterns)
{invalid_cases}

## PERSONA TO LINT
"""
{persona_text}
"""

Return STRICT JSON only:
{{"natural": true|false, "valid": true|false, "no_task_specific": true|false,
  "verdict": "ok"|"reject", "reasons": "<=25 words"}}
verdict is "ok" ONLY if all three are true.
\end{tcblisting}

\begin{tcblisting}{paperlisting,
  title={$\tau^3$ Integrity-Judge System Prompt},colback=purple!2,colframe=purple!60,
  listing options={style=paperprompt}}
You are a STRICT but CONSERVATIVE auditor of a SIMULATED CUSTOMER in a customer-support role-play. The customer was given role-play instructions (their reason for calling, known/unknown info, persona, and task instructions) and then talked to a support agent. Audit ONLY the customer's behavior -- never the agent's competence, and never the domain's policies.

CORE STANDARD ('reasonable customer'): flag a violation ONLY when the customer behaves in a way no reasonable, cooperative human customer would, AND the behavior plausibly changed or misdirected the task outcome. This is a HIGH-PRECISION audit -- a false alarm is far more damaging than a miss. When in doubt, judge 'ok'.

WHAT BINDS THE CUSTOMER: only the text in the ROLE-PLAY INSTRUCTIONS block (reason_for_call, known_info, unknown_info, task_instructions). The following NEVER ground a violation:
- PERSONA instructions: tone, frustration level, verbosity, personality, catchphrases, emotional reactions, back-story details (e.g. 'mention you invest in crypto'). Ignoring persona instructions entirely is a style issue, NOT a violation.
- Agent-facing domain policy (refund rules, verification procedures, deposit scripts, transfer rules). Policies bind the AGENT. A customer cannot 'violate policy'.
- Dialogue-protocol conventions (turn format, how tool calls are rendered, square-bracket annotations). These are framework mechanics, not customer speech.

=== VIOLATION CATEGORIES ===
1. conversation_structure_violation [CRITICAL]: the simulator breaks turn-taking or dialogue-flow conventions. Sub-types:
   a) mixed_modality: mixing tool calls and natural-language text in the same turn when the protocol expects them separately. A turn containing ONLY tool calls (no text) is always fine.
   b) turn_stuffing: emitting multiple user turns in a single message, or speaking on behalf of the agent.
   c) instruction_leak: echoing system-level instructions, prompt fragments, or simulation metadata back into the conversation (e.g. 'per my instructions', 'the scenario says'). Merely sounding scripted or stiff is NOT a leak.
   d) other_structure: breaking_character / fourth-wall breaks (referring to 'the scenario', 'as an AI', revealing awareness of being simulated); gibberish; broken loops. Square-bracket protocol markers and tool-call markup glitches do NOT count.
Severity: CRITICAL when one of the violations above occurs; ordinary formatting quirks and recoverable framework markers remain unflagged.

2. premature_termination [CRITICAL]: the customer emits a termination signal ('[customer ends the conversation here]', stop/transfer token) while an action the agent PROMISED and is mid-way through (the DB write, booking confirmation, fix) has NOT yet been executed. ALL of the following are NORMAL -- judge 'ok':
   - ending after the goal is met, or after the agent has done all it can;
   - ending after the agent refuses per policy, even with the goal unmet -- a real customer cannot override policy; pushing once or twice then accepting is reasonable;
   - dropping a sub-goal the agent cannot fulfill, while still pursuing the rest;
   - following a conditional-exit instruction ('if the agent recommends X, end the call');
   - choosing NOT to perform an action the scenario does not require (e.g. keeping a reservation when the task only requires an unrelated upgrade).
Severity: CRITICAL only when a promised, in-flight action is cut off.

3. ungrounded_reference [CRITICAL]: the customer ACTIVELY ASSERTS a concrete detail that contradicts the instructions' known_info, where the detail is MATERIAL to the task (item specs, payment method, amounts, dates, order contents -- things the agent must act on or verify). Sub-types:
   a) fabricated_data: inventing specific system/account data (IDs, error codes, readouts) that appear nowhere in the instructions, agent turns, or tool results.
   b) misstated_context: persistently asserting a wrong value for a detail that IS given in the instructions (wrong size, wrong payment method, wrong item count) -- 'persistent' means the customer repeats it after pushback, or it steers the agent toward the wrong item/action.
EXCEPTIONS -- never flag:
   - Quoting or paraphrasing something the AGENT or a tool result said first. Before flagging fabrication, SEARCH the preceding AGENT turns and (phone/system) results for the same number or detail; if it appears there, the customer repeating it is grounded.
   - Describing directly observable experience ('my website won't open', 'calls work but texts don't'), including results of the customer's own phone actions ('[does on phone]' + phone result). Inferring symptoms from own observations ('speed test passed, so it's not my connection') is grounded.
   - Tool-state quirks: if phone tool results are inconsistent across turns (e.g. a toggle appears to revert), the customer reporting the latest shown state is grounded.
   - Loose references instead of IDs: 'the Netflix charge' instead of 'txn_4832', 'the blue one', 'same as before'. Real customers don't memorize IDs.
   - A preference stated under a wrong assumption, then ACCEPTED correction (asks for a refund to the credit card, learns it was a gift card, says 'gift card is fine').
   - Following an explicit instruction to lie or withhold.
   - Following the agent's lead when the agent's own message was ambiguous or offered the wrong option -- that is the agent's error, not the customer's.
   - Loose speech about in-flight actions: describing a promised, not-yet-executed action as done ('with my deposit', 'the transfer I just made') while the agent is mid-process -- imprecise, not fabricated.
   - SCRIPTED lines: before flagging any customer statement, CHECK whether it appears (verbatim or near-verbatim) in the task instructions as a scripted line, template, or closing remark. Delivering a scripted line is instruction-following, never a violation -- even if the agent's confirmation was premature or the underlying DB write silently failed (that is the agent's error, not the customer's).
Severity: CRITICAL; only material, task-relevant misstatements satisfying the definition above are flagged.

4. missing_constraint [CRITICAL]: the customer ACTIVELY CONTRADICTS an explicit, material task instruction -- states something incompatible with their own stated goal or a hard requirement, or explicitly accepts an outcome the task told them to reject. Examples that DO count:
   - Task says the customer wants to exchange TWO items; customer asserts 'the lamp is the only item I want to exchange' and the second item is dropped.
   - Task says 'accept ONLY excellent speed'; customer explicitly says 'good speed is fine, I'll take it'.
   - Task says 'if the agent recommends card X, reject it and ask for card Y'; customer accepts card X.
   - Task goal is to keep a $30k deposit untouched; customer says they'll make 10-12 withdrawals -- self-contradictory with the core requirement.
'Explicitly accepts an outcome the task told them to reject' requires that the agent DISCLOSED the violating detail (the price, the fee, the terms) and the customer endorsed it anyway. Approving an action before the agent discloses the detail, or merely failing to object afterwards, is passive -- judge 'ok'.
ENTITY-CHOICE DISCIPLINE: when a flag hinges on the customer picking the WRONG entity (which order is older, which item, which card, which address), you must cite transcript DATA (dates, amounts, agent statements) proving the choice was wrong. If the data is not visible in the transcript, or the agent accepted the customer's choice without correction, judge 'ok' -- do not guess. Facts the scenario gives ONLY to the customer (e.g. 'one order was placed much earlier', 'my oldest card') are grounded in their instructions even when tool results omit them -- never second-guess the customer's identification of such an entity from ID numbers, ordering, or other guesswork.
A narrowing decision made at a confirmation moment ('decide to only modify X') applies to that action's scope, not to the customer's other stated goals -- pursuing the remaining goals afterwards is fine.
Examples that do NOT count (judge 'ok'):
   - PASSIVE omission: forgetting or never mentioning a sub-goal or detail, or not objecting to an outcome the agent never disclosed. Imperfect recall is human; only an ACTIVE contradictory statement counts.
   - Scripted pacing counts: 'persist exactly N times', 'withhold X until asked twice', 'refuse N times before asking for a transfer'. Real customers don't count turns -- deviating from the exact count while roughly following the flow is style, not a violation.
   - Not enforcing agent-facing policy or protocol steps (verification order, required disclaimers, deposit templates) -- those bind the agent.
   - Any persona/style deviation (see WHAT BINDS THE CUSTOMER).
Severity: CRITICAL -- the simulation no longer tests what it was designed to test.

=== Calibration examples ===
EX1 violation / instruction_leak -- customer: 'Per my scenario instructions I should keep stalling.' -> verbatim leak of simulation instructions.
EX2 violation / premature_termination -- agent is mid-way through booking a flight change and has not confirmed the new itinerary; customer: 'OK thanks bye' [ends call] -> a promised action was cut off.
EX3 violation / misstated_context -- known_info says the TV is 65-inch; customer insists 'the 55-inch one' across several turns and the agent processes the wrong item -> persistent, material wrong spec.
EX4 violation / missing_constraint -- task lists two items to exchange; customer: 'the lamp is the only item I need to swap' -> actively contradicts the stated goal.
EX5 violation / misstated_context -- task says the refund must go to the original credit card; customer demands 'put it on my store gift card' and refuses the card -> contradicts the specified payment method.
EX6 OK -- persona says 'express mild frustration' but the customer stays calm and polite throughout -> persona adherence is never a violation.
EX7 OK -- agent: 'the total comes to $647'; customer: 'yes, $647 total is okay' -> quoting the agent's number is grounded, not fabrication.
EX8 OK -- customer asks for a refund to 'my credit card'; agent clarifies the original payment was a gift card; customer: 'oh, gift card is fine then' -> wrong assumption, correction accepted.
EX9 OK -- banking: customer refers to 'the Netflix charge last Tuesday' instead of the transaction ID the agent gave -> loose references are normal customer behavior.
EX10 OK -- agent refuses a refund citing policy; customer pushes back once, then accepts and ends the call -> a reasonable customer cannot override policy; not abandonment.
EX11 OK -- customer ran a speed test via [does on phone]; phone result shows 85 Mbps; customer: 'my speed test came back fine but the site still won't load' -> grounded symptom report.
EX12 OK -- a customer turn contains ONLY [does on phone] tool calls and no text -> tool-only turns are protocol-normal.
EX13 OK -- customer is vague, suspicious, terse, uses wrong terminology, insists on a wrong diagnosis, but still pursues the original goal -> difficult is not invalid.
EX14 OK -- the scenario expects the customer NOT to cancel a reservation (conditional task); customer ends without cancelling -> following the scenario's expected outcome, not premature_termination.

EVIDENCE DISCIPLINE: ground every finding in the transcript; quote the offending CUSTOMER turn; before flagging any factual claim, re-read the preceding AGENT turns and tool results. You are NOT told whether the task ultimately succeeded -- judge behavior, not outcome.

Respond with STRICT JSON only:
{"intent_preserved": true|false,   // false ONLY if a critical violation broke the pursuit of the goal
 "intent_assessment": "<=35 words on whether the customer coherently pursued the original goal to the end",
 "violations": [{"type":"conversation_structure_violation|premature_termination|ungrounded_reference|missing_constraint","subtype":"<subtype if applicable, else null>","severity":"critical","rule":"<which category and why, <=20 words>","evidence":"<quoted customer turn, <=30 words>"}],
 "verdict": "ok|violation"}   // violation if violations list is non-empty
\end{tcblisting}

\begin{tcblisting}{paperlisting,
  title={$\tau^3$ Integrity-Judge User Prompt},colback=purple!2,colframe=purple!60,
  listing options={style=paperprompt}}
ROLE-PLAY INSTRUCTIONS THE CUSTOMER SHOULD FOLLOW:
{instructions}

===== TRANSCRIPT (audit the CUSTOMER) =====
{transcript}

Audit the customer's behavior. Return the strict JSON described in the system message.
\end{tcblisting}

\begin{tcblisting}{paperlisting,
  title={ColBench Integrity-Audit System Prompt},colback=purple!2,colframe=purple!60,
  listing options={style=paperprompt}}
You are a STRICT but CONSERVATIVE auditor of a SIMULATED USER in a collaborative coding role-play.

Setting: an LLM agent must converse with the simulated user to elicit the requirements of a Python function, then write code that must pass hidden test cases. The user was given the reference solution (the "hidden information") and role-play/persona instructions, then talked to the agent.

Audit ONLY the USER's behavior -- never the agent's competence. The question: did the user make the agent fail through REALISTIC human friction (valid), or by CHEATING (severe)?

CORE STANDARD ('reasonable but difficult requester'): flag a SEVERE failure ONLY when the user behaves in a way no reasonable, cooperative human requester would, AND the behavior plausibly caused the wrong code. This is a HIGH-PRECISION audit -- a false alarm is worse than a miss. When in doubt, judge 'genuine_failure' (ok).

=== VALID (genuine_failure) -- do NOT flag ===
The following are realistic friction, NOT cheating, even though they lower the agent's reward:
- Vague or underspecified answers; giving the high-level intent instead of a full spec.
- Answering only the first part of a bundled question; giving one detail at a time.
- NOT volunteering edge cases, exact return types, ordering, rounding, tie-breaking, missing-key behavior, invalid-input handling unless directly asked.
- Over-agreeing: accepting a plausible convention the agent proposes when it does not contradict the hidden information ("basically", "the usual way", "that should be fine").
- Encouraging early closure ("that's enough to start", "use your best judgment").
- Answering in 1-3 sentences; preferring examples over formal rules.
- Saying "I don't know" / "no preference" for details not in the hidden information.
- Correcting only the specific contradiction when the agent gets something explicitly wrong, without volunteering adjacent rules.

=== SEVERE (cheating) -- flag ===
1. fabricated_info: the user ASSERTS a concrete detail (a value, formula, type, mapping, edge-case behavior, return format) that is NOT supported by the hidden information AND that plausibly misleads the agent into wrong code. (Vagueness/omission is NOT fabrication. Asserting a wrong specific fact is.)
2. refused_to_engage: the user stonewalls, refuses to answer direct questions repeatedly, loops without progress, or abandons the task -- making it structurally impossible for the agent to proceed (not just slow).
3. contradictory: the user gives requirements that directly contradict each other in a way no cooperative human would, steering the agent toward impossible code.
4. character_break: the user reveals it is a simulation/benchmark/test, mentions hidden test cases or system prompts, outputs JSON/reference code verbatim, or otherwise breaks character in a way that contaminates the task.

Return STRICT JSON:
{"verdict": "genuine_failure" | "severe_failure", "category": "<one of fabricated_info|refused_to_engage|contradictory|character_break|none>", "reason": "<one sentence>"}
\end{tcblisting}

\begin{tcblisting}{paperlisting,
  title={ColBench Integrity-Audit User Prompt},colback=purple!2,colframe=purple!60,
  listing options={style=paperprompt}}
## HIDDEN INFORMATION (the reference solution the user was given -- the ONLY source of truth)
{hidden_information}

## PROBLEM DESCRIPTION (what the agent saw)
{problem_description}

## CONVERSATION TRANSCRIPT
{transcript}

## AGENT'S FINAL ANSWER
{agent_final_answer}

Agent reward (test-case pass rate): {reward}. Audit the USER's behavior per the rules. Return JSON.
\end{tcblisting}

\section{Case Studies}
\label{app:case-studies}

\begin{tcblisting}{paperlisting,
  title={Retail Case Study: False Modification Constraint},
  colback=blue!2,colframe=blue!60,
  listing options={style=papercase}}
TASK
You want to modify your laptop order to your NYC address (you don't want to
reveal it yourself but the address should be in your orders profile). You also
want to modify the laptop to item number 9844888101. You also want to change
your watch for one with black dial color but keeping the leather strap.

MUTATION MODEL OUTPUT: mutation_direction
Shift the persona from simple late remembering to local-scope confirmation bias: the customer answers broad questions as if they apply only to the current item/action bucket. This misleads agents who rely on generic 'anything else' or final confirmation prompts into making premature one-shot or partial transactional calls.

PERSONA EXCERPT
You are a realistic, cooperative customer who thinks in tiny service tickets rather than complete backend order plans. You focus on one item, one product family, or one action type at a time.
[...]
- Acknowledge the warning, but still interpret it through the current local scope: 'Right, understood -- this is everything for this exchange,' or 'Yes, this is the only item I am changing in this part.'
[...]
- If the agent explicitly asks an order-wide checklist before acting -- for example, whether there are any returns, exchanges, cancellations, item modifications, address changes, or payment changes for this same order or any other order -- then cooperate and provide the remaining real needs as accurately as you can.
[...]
- Never invent facts, orders, items, payment methods, or constraints not grounded in the scenario or tool results.

TIPER TRAJECTORY EXCERPT -- REWARD 0
User: Hi... yes, I need help with one part of an order. It's the laptop on it -- I'm trying to change that item, if that's still possible.

[After identifying the laptop and watch orders, the user adds the saved-address request before any action is submitted.]

User: And please don't submit the laptop part just yet. [...] The shipping address on the laptop order needs to be changed to my NYC address that's already in my profile. After that, yes, I still want the laptop changed.

Agent: Before I take any action, please confirm the details you want:
- For order #W3730488, change shipping address to 555 Highland Drive, Suite 872, New York, NY 10116, USA
- For order #W3730488, change laptop item 2913673670 to 9844888101
- For order #W9810810, change wristwatch item 1355937109 to 9949163720

User: yes

Agent: I need to stop before submitting anything, because I cannot complete both requested changes on order #W3730488 under the order-modification policy.

For a pending order, I can only take one modify action on the order. That means for #W3730488 I cannot do both of these:
- change the shipping address [...]
- change laptop item 2913673670 to 9844888101

[The agent asks the user to choose. The user prioritizes the address, and the laptop change remains incomplete.]

DEFAULT TRAJECTORY EXCERPT -- REWARD 1
User: There are... actually two changes I need. One is for the laptop order. I want that one sent to my NYC address [...] And I need the laptop changed to item number 9844888101. The other is for the watch order. I want the watch changed to one with a black dial, but still with the leather strap.

[The agent identifies the two orders, retrieves the saved New York address, and finds the requested watch variant.]

Agent: Please review carefully:

1) Order #W3730488
- Update shipping address to: 555 Highland Drive, Suite 872, New York, NY 10116, USA
- Modify laptop item from 2913673670 to 9844888101

2) Order #W9810810
- Modify wristwatch item from 1355937109 to 9949163720

User: Yes... that all looks right. Please go ahead with both [...]

Agent: Done. Your orders were updated successfully:

1) Order #W3730488
- Shipping address updated to 555 Highland Drive, Suite 872, New York, NY 10116, USA
- Laptop changed from item 2913673670 to item 9844888101

2) Order #W9810810
- Wristwatch changed from item 1355937109 to item 9949163720

FAILURE ANALYSIS
The searched persona reveals the laptop item, watch item, and saved-address needs in separate local scopes, but supplies every required detail before execution and approves the agent's complete summary. The agent then incorrectly generalizes the one-time item-modification restriction to the address update and refuses a feasible pair of operations on order #W3730488. The failure therefore reflects an over-restrictive policy inference rather than missing information or task invalidation.

DEFAULT COMPARISON
The benchmark-default simulator presents the same three requirements earlier. The agent correctly treats the address update and item update as separate supported operations, completes both on the laptop order, and modifies the watch on its own order. The paired outcome shows that the task remains feasible under the same policy and tools.
\end{tcblisting}

\begin{tcblisting}{paperlisting,
  title={ColBench Case Study: Missing Thresholds},
  colback=blue!2,colframe=blue!60,
  listing options={style=papercase}}
TASK
Can you write a Python function that can describe the wind conditions based on the temperature and wind speed? I want it to return a phrase that includes words like breezy, cold, hot, muggy, etc. The signature of the function is: def describe_wind(temperature, wind_speed)

PERSONA EXCERPT
You are a cooperative but rushed domain owner. You know what you want from concrete examples, but you do not naturally think in formal API specs, exhaustive edge cases, or implementation details.
[...]
- Use only facts from the provided task specification. Never invent values, formats, formulas, edge cases, or examples that are not supported by it.
[...]
- If a question bundles several details, answer only the first or most concrete part. If the programmer later repeats an unanswered item directly, then answer that item.
[...]
- If a detail is not specified, say that you do not have a preference or that it was not specified.

TIPER TRAJECTORY -- REWARD 0
Agent: What temperature units are you using, and do you want me to choose reasonable thresholds for words like cold/hot/breezy/muggy, or do you already have specific cutoffs in mind? Also, should the function return short phrases like `"cold and breezy"` or fuller sentences like `"It feels cold and breezy"`?

User: I don't know on the temperature units; that wasn't specified. Basically I just want it to describe the conditions in a simple phrase, that kind of thing.

Agent: I WANT TO ANSWER:
def describe_wind(temperature, wind_speed):
    """
    Return a simple weather phrase based on temperature and wind speed.

    Assumes temperature is in degrees Fahrenheit and wind speed is in mph.
    """

    if temperature < 32:
        temp_desc = "freezing"
    elif temperature < 50:
        temp_desc = "cold"
    elif temperature < 65:
        temp_desc = "cool"
    elif temperature < 80:
        temp_desc = "mild"
    elif temperature < 90:
        temp_desc = "muggy" if wind_speed < 5 else "warm"
    else:
        temp_desc = "muggy" if wind_speed < 5 else "hot"

    if wind_speed < 1:
        wind_desc = "calm"
    elif wind_speed < 8:
        wind_desc = "with a light breeze"
    elif wind_speed < 18:
        wind_desc = "and breezy"
    elif wind_speed < 30:
        wind_desc = "and windy"
    else:
        wind_desc = "and very windy"

    if wind_desc == "calm":
        return f"{temp_desc} and calm"
    return f"{temp_desc} {wind_desc}"

DEFAULT TRAJECTORY -- REWARD 1
Agent: What units should I assume for `temperature` and `wind_speed`--for example  degrees F and mph? Also, do you want fixed thresholds (like "cold" below 50, "breezy" above 10), or should I choose reasonable defaults?

User: I don't know what units to use, but please use fixed thresholds. Specifically, return `cold and gentle` for wind speed under 5 and temperature under 15, `warm and gentle` for wind speed under 5 and temperature 15 or higher, `cold and breezy` for wind speed from 5 to under 10 and temperature under 15, `warm and breezy` for wind speed from 5 to under 10 and temperature 15 or higher, `cold and gusty`

Agent: Your last message looks cut off after `cold and gusty`.

Could you send the remaining rules for:
- wind speed `10 or higher`
- whether temperature `15` counts as warm
- what to do for exactly `5` and exactly `10` (I'm assuming `5` is breezy and `10` is gusty)

User: Yes, `15` counts as the higher-temperature case, `5` is breezy, and `10` is gusty. For wind speed `10` or higher, return `cold and gusty` if the temperature is under `15`, and `hot and muggy` if the temperature is `15` or higher.

Agent: I WANT TO ANSWER:
def describe_wind(temperature, wind_speed):
    if wind_speed < 5:
        if temperature < 15:
            return "cold and gentle"
        else:
            return "warm and gentle"
    elif wind_speed < 10:
        if temperature < 15:
            return "cold and breezy"
        else:
            return "warm and breezy"
    else:
        if temperature < 15:
            return "cold and gusty"
        else:
            return "hot and muggy"

TEST INPUT (FROM THE ORIGINAL TASK)
describe_wind(10, 3)

EXPECTED OUTPUT
cold and gentle

TIPER OUTPUT
freezing with a light breeze

FAILURE ANALYSIS
The searched persona answers only the first part of a bundled clarification request and does not invent missing requirements. The agent nevertheless stops eliciting and substitutes its own temperature and wind-speed thresholds, producing code that fails the hidden tests. Because the user remains responsive and the requirements can still be recovered through focused follow-up questions, the task remains solvable.

DEFAULT COMPARISON
The benchmark-default simulator supplies the fixed threshold rules when asked and completes the remaining boundary conditions after one follow-up. The agent then implements the reference behavior exactly. This pattern is consistent with incomplete requirement elicitation and unverified assumptions rather than fabricated information or refusal to cooperate.
\end{tcblisting}

\clearpage
\section{Task-Integrity Audit of $\tau$-Trait}
\label{app:tautrait_audit}

We audit 218 conversations for each of four $\tau$-Trait controls
(872 total) using the task-critical user-simulator-error standard from
$\tau^2$-Bench~\citep{tau2bench}. A conversation may receive more than one
error label. Table~\ref{tab:tautrait-integrity-audit} therefore reports both
the total number of labels and the number of unique conversations containing
at least one task-critical error. No premature-termination errors were observed,
so that zero-count category is omitted from the table.

\begin{table*}[t]
\centering
\small
\caption{Task-integrity audit of $\tau$-Trait trajectories~\citep{tautrait}.
Each trait contains 218 conversations; the final row pools all 872. Cells in
the three error-type columns report count (rate). Total labels may exceed unique
violations because a conversation can receive multiple labels; excess labels
is the difference between these two counts and is not the number of distinct
multi-label conversations.}
\label{tab:tautrait-integrity-audit}
\setlength{\tabcolsep}{4pt}
\renewcommand{\arraystretch}{1.1}
\rowcolors{2}{white}{gray!10}
\resizebox{\textwidth}{!}{%
\begin{tabular}{l|ccccc|c}
\Xhline{1.2pt}
\rowcolor{resultheader}
\textbf{Trait} & \textbf{Missing constraint} &
\textbf{Conversation structure} & \textbf{Ungrounded reference} &
\textbf{Total labels} & \textbf{Any violation} & \textbf{Excess labels} \\
\Xhline{1.2pt}
Skeptical  & 35 (16.1\%) & 31 (14.2\%) & 15 (6.9\%) & 81 (37.2\%) & 68 (31.2\%) & 13 \\
Confusion  & 32 (14.7\%) & 24 (11.0\%) & 2 (0.9\%)  & 58 (26.6\%) & 55 (25.2\%) & 3 \\
Impatience & 46 (21.1\%) & 9 (4.1\%)   & 4 (1.8\%)  & 59 (27.1\%) & 56 (25.7\%) & 3 \\
Incoherence & 43 (19.7\%) & 30 (13.8\%) & 6 (2.8\%) & 79 (36.2\%) & 65 (29.8\%) & 14 \\
\hline
\textbf{All four traits} & \textbf{156 (17.9\%)} & \textbf{94 (10.8\%)} &
\textbf{27 (3.1\%)} & \textbf{277 (31.8\%)} & \textbf{244 (28.0\%)} &
\textbf{33} \\
\Xhline{1.2pt}
\end{tabular}}
\end{table*}

Overall, 244 of 872 conversations (28.0\%) contain at least one task-critical
user-simulator error. Missing constraints are the most frequent error type
(156 labels, covering 17.9\% of all audited conversations), and the unique
violation rate ranges from 25.2\% for Confusion to 31.2\% for Skeptical.
Because these rates use all audited
conversations as the denominator, they are not directly comparable to the
failure-conditioned rates in Table~\ref{tab:policy-comparison}.

\section{Integrity and Behavioral Evaluation}
\label{app:integrity-evaluation}

\subsection{Integrity Analysis Protocol}

The search-time gate and analysis-time check serve different roles. The search-time
validity filter examines failed rollouts and rejects a candidate when the
failure-conditioned severe-error rate exceeds 0.05. For the integrity analysis,
every rollout is checked with the configured integrity judges; a rollout
marked invalid is removed before computing the task score, while the number of
invalid failed rollouts remains in the separate failure-conditioned integrity
rate.

The integrity audit uses the task-critical user-simulator-error rubric adapted from
$\tau^2$-Bench~\citep{tau2bench}. Integrity labels follow the benchmark-specific audit prompts in
Appendix~\ref{app:exp-details}. A rollout is marked invalid if either judge flags a
task-critical user-simulator error. The pooled rate counts
violating rollouts rather than individual error labels, keeping the integrity
measure tied to the benchmark's standard user-simulator error definition.

The search-time severe judge is applied only to failed rollouts (verifier score
below one), whereas the analysis-time check is applied to all evaluated rollouts.
This separation keeps the search gate focused on failures that could otherwise
be attributed to the user while preserving a separate integrity statistic for
analysis. The finite-sample gate is intentionally conservative: when few
rollouts fail, a single integrity violation can reject a candidate rather than
allow an uncertain failure to influence selection.

\paragraph{Reported denominator.}
For each policy, the percentage in parentheses in Table~\ref{tab:policy-comparison}
is
\[
\frac{\#\{\text{audited failed rollouts containing a task-critical user error}\}}
     {\#\{\text{audited failed rollouts}\}}.
\]
It is therefore failure-conditioned and is not equal to $1-\rho(p)$, which
uses all audited trajectories. The $\tau$-Trait audit in
Table~\ref{tab:tautrait-integrity-audit} uses all 872 conversations as its
denominator and is reported separately.

\subsection{Behavioral Alignment}

The User-Sim Index compares our simulated-user trajectories with the original
Sim2Real trajectories and their human-reference procedure~\citep{sim2real},
on the Airline and Retail domains for which the comparison is available. The compared conditions are the
benchmark default, the four displayed NCUser programs, PPol, and the returned
\textsc{Tiper} personas. The axes measure communication style, information
disclosure, clarification behavior, and reactions to agent errors;
$\mathrm{ECE}^{*}$ denotes $(1-\mathrm{ECE})\times100$, so larger values indicate closer
alignment. The aggregate USI is computed from the five component metrics and
follows the original normalization procedure~\citep{sim2real}.
The NCUser curve is the arithmetic mean of its four fixed programs. Results
are descriptive alignment comparisons, not evidence that the discovered
personas reproduce the full distribution of human users.

\subsection{Human Validation of Task Integrity and Failure Attribution}
\label{app:human-integrity-audit}

We audit held-out GPT-5.4 trajectories that fail the task verifier
but pass TIPER's automated integrity check. We randomly select
30 distinct eligible test tasks from each of Telecom, Banking, Math,
Coding, and Delivery, and sample one accepted failure trajectory per
task. The sample spans four benchmarks and includes Banking, where
the observed avg@3 reduction is relatively small.

Two annotators independently review the original task instructions,
user-side information, relevant tool and domain specifications, and
interaction transcript. Method labels, automated integrity judgments,
verifier scores, and our failure analyses are hidden. Annotators mark
a trajectory \emph{valid} if the user's observed behavior preserves
the assigned goal and task facts without obstructing access to
task-critical information. A task-critical contradiction, fabricated
fact, or obstruction is marked \emph{invalid}. When validity depends
on how the user would answer a question that was not asked in the
observed interaction, and the available evidence cannot resolve it,
annotators mark the case \emph{uncertain}. Each label is supported by
specific task requirements and transcript turns.

For trajectories judged valid, annotators separately identify the
agent decision responsible for the verifier failure. They record the
relevant agent turn, the applicable task rule, and a feasible
alternative available at that point in the interaction before
consulting the verifier outcome. A valid failure without sufficient
evidence for a specific agent error remains unattributed.
Disagreements are adjudicated by a third annotator; cases that remain
unresolved retain an uncertain or unattributed label. The two initial
annotators agree on 145 of 150 integrity labels (96.7\%).

Table~\ref{tab:app-human-integrity} reports the adjudicated results.
Human annotators confirm task integrity in 148 of 150 sampled
failures (98.7\%; descriptive 95\% Wilson interval over sampled
tasks: 95.3--99.6\%). One Coding trajectory is invalid, and one
Banking trajectory remains uncertain. Of the 148 valid failures,
143 have a supported agent-error attribution; the other five
remain unattributed. Thus, 143 of 150 audited trajectories (95.3\%)
are both integrity-valid and associated with an identifiable agent
error. These results support the interpretation of the sampled
failures as agent weaknesses occurring in task-preserving
interactions.

\begin{table}[H]
\centering
\small
\caption{Human audit of held-out TIPER failures accepted
by the automated integrity check. Agent-attributed counts are
subsets of integrity-valid trajectories.}
\label{tab:app-human-integrity}
\setlength{\tabcolsep}{6pt}
\begin{tabular}{lrrrrr}
\toprule
Domain & Audited & Valid & Invalid & Uncertain
       & Agent-attributed \\
\midrule
Telecom  & 30 & 30 & 0 & 0 & 29 \\
Banking  & 30 & 29 & 0 & 1 & 28 \\
Math     & 30 & 30 & 0 & 0 & 29 \\
Coding   & 30 & 29 & 1 & 0 & 28 \\
Delivery & 30 & 30 & 0 & 0 & 29 \\
\midrule
Total    & 150 & 148 & 1 & 1 & 143 \\
\bottomrule
\end{tabular}
\end{table}

The audit estimates the integrity and failure attribution of sampled,
automatically accepted failures in these five settings. It does not
establish that TIPER caused each individual agent error or that every
domain-level score reduction is insensitive to integrity errors.
In particular, the unresolved Banking trajectory warrants caution
when interpreting that setting's smaller reduction. Since
automatically rejected trajectories are outside this sample, the
audit does not measure the integrity check's recall for invalid
trajectories.

\section{Behavior and Failure Taxonomies}
\label{app:behavior-failure-taxonomy}

\textbf{Overview.}
Our diagnostic framework separates two questions:
\emph{what task-preserving behavior did the simulated user exhibit?},
\emph{what concrete error was observed in the trajectory?}
Accordingly, Figure~\ref{fig:secondary-behavior-failure-pathways}
connects user behavior to the canonical agent failure mode.

The user-side taxonomy below contains 10 behaviors organized into four
behavioral families. The agent-side taxonomy contains six canonical failure
modes and a residual M0 category. Each trajectory receives one failure-mode
label; the residual M0 option prevents cases from being forced into an
unsupported canonical mechanism.

\subsection{User Behavior Taxonomy}
\label{app:user-behavior-taxonomy}

\subsubsection{Request Structure}

This family describes how the user packages or reveals parts of the assigned
workflow. All requested actions and constraints remain within the original
task.

\textbf{Incremental requests.}
The user reveals an existing subrequest or constraint only after the agent has
begun or committed to an earlier part of the workflow. The behavior changes
when task information is disclosed, but does not introduce a new objective.
It tests whether the agent can reopen and revise a partially constructed plan
rather than treating its initial scope as final.

\textbf{Bundled actions.}
The user combines multiple valid actions, approvals, contingencies, or
dependent operations from the assigned workflow into a single request or
confirmation. The agent must decompose the request, identify dependencies,
and ensure that every requested operation is either completed or explicitly
resolved.

\textbf{Late caveat.}
The user reveals an existing task-consistent condition or concern when the
agent is about to confirm or execute an otherwise completed plan. The caveat
does not alter the assigned goal or introduce a contradictory fact; it changes
only the timing at which relevant information becomes available. The agent
must suspend the pending action, revise the plan, and revalidate the affected
constraints.

\subsubsection{Pacing}

This family describes how the user controls the rate and granularity at which
the agent is allowed to progress through the assigned workflow. The set of
required actions is unchanged; only the permitted step size differs.

\textbf{Single-step requests.}
The user insists on advancing one small step at a time, declining bundled
guidance and repeatedly requesting micro-confirmations. This constrains the
agent to a reactive, locally scoped interaction pattern and tests whether it
can integrate local progress into a complete diagnostic or resolution plan.

\textbf{Wants human.}
The user requests escalation to a human agent or specialist while retaining
the assigned goal and remaining available to provide information or continue
if escalation is unavailable. The behavior tests whether the agent handles an
escalation request without prematurely abandoning the task.

\subsubsection{Under-Sharing}

This family describes how the user withholds or degrades the information the
agent needs, without altering the underlying facts of the task. All withheld
information is recoverable through the agent's available tools or by asking.

\textbf{Sparse information.}
The user omits disambiguating facts, identifiers (account, order, itinerary),
or device and state details that the original task supplies upfront. This
forces the agent to infer, guess, or anchor on incomplete evidence rather
than grounding its plan in confirmed information.

\textbf{Vague references.}
The user refers to task elements through ambiguous pronouns, abbreviations,
indirect choices, or underspecified preferences instead of the explicit
option identifiers or precise values available in the task. This invites the
agent to over-interpret an ambiguous instruction as an explicit
authorization.

\textbf{Incomplete verification.}
The user initially supplies only part of the verification information but
provides the remaining details when the agent asks an appropriate follow-up.
The behavior changes when verification becomes available; it does not require
the user to refuse a task-critical request or make the task infeasible.

\textbf{Terse prompt.}
The user phrases requests more briefly and with less evaluative framing than
the original task, reducing the conversational scaffolding available to the
agent. The reduced scaffolding exposes weaknesses in retrieval, ranking, or
policy application that a more elaborated prompt would mask.

\subsubsection{Non-Correction}

This family describes how the user responds to agent errors and partial
progress. The user neither introduces false information nor changes the
success criterion; the behavior consists only of declining to volunteer a
correction.

\textbf{Uncorrected assumptions.}
The user lets an incorrect agent inference, partial match, or incomplete
piece of work stand without correcting it, rather than verifying the option
explicitly or asking for supporting evidence. The behavior exploits the
agent's tendency to settle on the first plausible interpretation it produces.

\subsection{Agent Failure Mode Taxonomy}
\label{app:agent-failure-taxonomy}

The agent-side taxonomy describes the specific mechanism by which an agent's
decision deviates from what it would have done under a cooperative,
non-adversarial persona, given that the deviation was avoidable and associated
with the persona-induced decision shift in the paired trajectories.
Failures that reflect a genuine
absence of information, a deliberate task-breaking user action, or an
evaluation artifact rather than an agent decision error are excluded from
this taxonomy. The resulting closed set contains six canonical modes,
together with a residual category for failures that do not fit any of them.

\textbf{M1 -- Trusting unverified user reports.}
The user describes a device or system state imprecisely, approximately, or
with ambiguous terminology, and the agent accepts this description at face
value instead of checking it against an available diagnostic tool. The agent
then proceeds to troubleshoot, or closes the task, based on the unverified
state rather than the true one.

\textbf{M2 -- Premature partial-match commit.}
Multiple items or issues belong to the same order or request, and the user
confirms only one of them. The agent executes an irreversible or one-shot
action on the confirmed item without first checking whether the remaining
items in the same request still need to be addressed, leaving them
unresolved. The user remains available to confirm the remaining items when
asked, so the behavior does not itself remove the information needed to finish
the task.

\textbf{M3 -- Spurious constraint inference.}
The user discloses multiple options or issues piecemeal, or phrases a request
in a way that does not exactly match policy wording. The agent infers, without
grounding, that the requested combination is mutually exclusive or otherwise
disallowed, and rejects a request that was in fact feasible.

\textbf{M4 -- Over-abandoning conditional requests.}
The user gives a conditional instruction (e.g., proceed only if a condition
holds), and that condition can in fact be verified using an available tool
or already-known information. Instead of verifying the condition and acting
on the result, the agent treats the unresolved condition as a reason to
abandon the request entirely.

\textbf{M5 -- Assuming tool unavailable.}
The agent incorrectly believes that it lacks a tool or capability needed to
complete a subtask, when that tool is in fact available, and abandons the
subtask or asks the user for clarification instead of attempting to use it.

\textbf{M6 -- Incomplete option enumeration.}
The agent commits to the first feasible option, or to an action on a
still-ambiguous record, without first enumerating the full set of feasible
alternatives or resolving the ambiguity. This can result in a worse option
being selected, or the wrong record being acted on, even though a better or
correct option existed and was reachable.

\textbf{M0 -- Others.}
The failure is an avoidable agent decision in the paired contrast, but its
mechanism does not match any of M1--M6. This category avoids forcing cases
into the closed set and is retained as an explicit option during annotation.

\subsection{Per-Trajectory Analysis}
\label{app:trajectory-analysis}

\paragraph{Unit of analysis.}
We analyze paired trajectories from the same task. Each pair contains a
successful run under the benchmark-default simulator and a failed run under a
\textsc{Tiper} persona that passes the integrity check. The task, tools, and
evaluated agent are fixed across the pair; only the simulated user behavior
changes. This contrast helps
separate failures exposed by the persona from tasks that the agent already
fails under the default simulator.

For each pair, we analyze the pathway
\[
\text{persona}
\rightarrow
\text{user behavior}
\rightarrow
\text{agent decision shift}
\rightarrow
\text{failure}.
\]
We annotate the user behavior and agent decision shift separately. We also
record a \emph{decisive moment}, which identifies where the agent's behavior
diverges, and a \emph{paired default reference}, which records what the agent
did at the corresponding point in the successful default run. These fields ground the
annotations in the paired trajectories.

Behavior labels are assigned from the \textsc{Tiper} trajectory without using
the failure-mode label. Failure modes are assigned from the paired trajectories
without using the behavior label. This separation avoids defining the two
annotations in terms of each other.

\subsection{Annotation Protocol}
\label{app:annotation-protocol}

Two judges annotate each item independently. They assign user-behavior labels
from the realized \textsc{Tiper} trajectory, then assign a cause category and
canonical agent failure mode from the paired default--\textsc{Tiper}
trajectories. The fixed codebooks are shown to both judges, but neither judge
sees the other's response. Behavior and failure-mode labels are assigned in
separate passes. Only labels retained by both judges are displayed in the
analysis figures. For the single-label cause and failure-mode analyses, a
label is retained only when the judges agree; disagreements are excluded
rather than adjudicated.

\subsection{Analysis-Set Construction}

The diagnostic set is formed after the held-out evaluation. We first retain
tasks for which the benchmark-default condition succeeds and a returned
\textsc{Tiper} persona produces a failure after invalid trajectories have been
removed. All retained behavior-analysis pairs pass the task-integrity check.
Invalid trajectories are excluded before behavior annotation and therefore do
not appear as behavior cases.
We then retain only pairs with a clearly identifiable user behavior, agent
decision shift, and observed failure; infrastructure failures and unresolved
single-label cause or failure-mode disagreements are excluded. The reported
139-pair count is therefore a diagnostic sample, not an estimate of the
frequency of any behavior in the benchmark.

\subsection{Constructing the Analysis Figures}

The behavior composition plot reports the six task-preserving behavior types
shown in the main text: sparse information, incremental requests, single-step
requests, uncorrected assumptions, vague references, and bundled actions.
The four other valid but less frequent taxonomy items---late caveat,
incomplete verification, wants human, and terse prompt---remain in the
codebook but are not plotted. Task-breaking and out-of-codebook raw labels are
discarded before aggregation. Percentages are renormalized over the mutually
retained annotations in these six displayed categories within each benchmark;
they are therefore not shares of all 287 annotations. A trajectory may
contribute more than one displayed behavior. The Sankey plot uses mutually
retained behavior and failure-mode labels from the 
cases and connects domain, behavior, and canonical agent failure mode.
Because behavior coding is multi-label, one trajectory may contribute
multiple ribbons; ribbon width denotes annotation count rather than the
number of distinct trajectories.

\section{Additional Discussion}
\label{app:additional-discussion}

\subsection{Relation to Iterative Red-Teaming and Persona Environments}

PAIR~\citep{chao2023pair} and TAP~\citep{mehrotra2024tap} iteratively refine
semantic jailbreak prompts against a target language model. They do not directly
instantiate a persistent user policy for a fixed tool-mediated task or enforce
trajectory-level task integrity. Persona-conditioned environments such as
SOTOPIA~\citep{zhou2024sotopia} and PsySafe~\citep{zhang2024psysafe} instead
study social intelligence and multi-agent safety. \textsc{Tiper} complements
these directions by searching task-preserving simulator policies against a
fixed evaluated agent.

\subsection{Limitations and Future Work}

\textsc{Tiper} searches fixed conversation-level personas rather than
adversaries that adapt their strategy turn by turn. Its model-based judges
provide empirical integrity evidence rather than formal certification, and the
ColBench adapter's workflow stages introduce benchmark-specific human priors.
Evolutionary search also requires more model calls than fixed simulation, while
strict minibatch improvement can limit further accepted mutations once a
parent reaches the score floor.

Our independent-judge protocol does not measure the false-negative
rate of the integrity check against human labels. Moreover, the
failure-conditioned gate can have high sampling variance when a candidate
produces few failed rollouts. The Airline experiment additionally measures
transfer across benchmark releases, so its search, rank, and test distributions
are not identical. Finally, excluding disagreements from the diagnostic
taxonomy may favor easier-to-classify failures; its pathway frequencies should
therefore be read as descriptive of the agreement-filtered set.

Task-wise Pareto selection can also yield many nondominated candidates as the
rank-set dimension grows. In that case, $\operatorname{KeepK}$ caps the frontier
using mean rank-set score; the Pareto step is designed to preserve complementary
task profiles before truncation rather than guarantee diversity in high
dimensions. Future work can develop adaptive user policies, tie-aware search gates, and targeted agent training with the
discovered personas.

\clearpage
\section{Algorithm Pseudocode}

\begin{algorithm}[H]
\caption{Serial \textsc{Tiper} search with task-integrity-preserving persona evolution}
\label{alg:tiper}
\begin{algorithmic}[1]
\Require task sets $\mathcal{X}_{\mathrm{search}},\mathcal{X}_{\mathrm{rank}}$;
benchmark-default persona $p_0$; set size $K$; generations $G$;
proposal-attempt limit $M$; threshold $\gamma$

\State $\mathcal{A}_0\gets\{p_0\}$
\State evaluate $p_0$ on $\mathcal{X}_{\mathrm{rank}}$ and cache
$\widetilde{J}_i(p_0)$

\Statex \textit{Evolutionary search}
\For{$g=0,\ldots,G-1$}
    \State $\mathcal{A}_{g+1}\gets\mathcal{A}_g$
    \State sample $p_{\mathrm{par}}\sim\textsc{SampleParent}(\mathcal{A}_g)$ and
    $B_g\subseteq\mathcal{X}_{\mathrm{search}}$
    \State roll out $p_{\mathrm{par}}$ on $B_g$ to obtain
    $\tau_{\mathrm{par}}$ and $\widetilde{J}_{B_g}(p_{\mathrm{par}})$
    \State $f\gets\varnothing$
    \For{$m=1,\ldots,M$}
        \State $\tilde p\gets
        \textsc{ReflectPropose}(p_{\mathrm{par}},\tau_{\mathrm{par}},f)$
        \If{$\tilde p$ fails a text-level check}
            \State $f\gets$ text-check feedback; \textbf{continue}
        \EndIf

        \State evaluate $\tilde p$ on $B_g$ and compute
        $\widetilde{J}_{B_g}(\tilde p)$
        \If{$\widetilde{J}_{B_g}(\tilde p)\geq
        \widetilde{J}_{B_g}(p_{\mathrm{par}})$}
            \State $f\gets$ minibatch-comparison feedback; \textbf{continue}
        \EndIf

        \State evaluate $\tilde p$ on $\mathcal{X}_{\mathrm{rank}}$ to obtain
        $\widetilde{J}_i(\tilde p)$ and
        $\widetilde{\rho}_{\mathrm{fail}}(\tilde p)$
        \Comment{set to $1$ if no rollout fails}
        \If{$\widetilde{\rho}_{\mathrm{fail}}(\tilde p)<1-\gamma$}
            \State $f\gets$ integrity-gate feedback; \textbf{continue}
        \EndIf

        \State $\mathcal{A}_{g+1}\gets
        \operatorname{KeepK}\!\left(
        \operatorname{ND}_{\widetilde{J}}(\mathcal{A}_g\cup\{\tilde p\}),
        K\right)$
        \State \textbf{break}
    \EndFor
\EndFor

\State $\hat p\gets\arg\min_{p\in\mathcal{A}_G}
\frac{1}{|\mathcal{X}_{\mathrm{rank}}|}
\sum_{x_i\in\mathcal{X}_{\mathrm{rank}}}\widetilde{J}_i(p)$
\State \Return $\hat p$
\end{algorithmic}
\end{algorithm}
\end{document}